%% file: ICML2020_UCRL3_FinalVersion.tex
\documentclass{article}

\usepackage{microtype}
\usepackage{graphicx}
\usepackage{subfigure}
\usepackage{booktabs}
\usepackage{hyperref}

\usepackage[accepted]{icml2020}
\usepackage{multicol}

\usepackage{amsmath,amsfonts,amssymb}
\usepackage{bm}

\usepackage{algorithm,algorithmic}
\renewcommand{\algorithmicrequire}{\textbf{Input:}}
\renewcommand{\algorithmicensure}{\textbf{Output:}}

\usepackage{graphicx}
\graphicspath{{.}{Figs/}}

\usepackage{color}
\definecolor{Gr}{rgb}{0.1, 0.6, 0}

\usepackage{natbib}

\usepackage{changepage}
\newenvironment{myproof}[1]{
	
	\begin{adjustwidth}{0.1cm}{0.1cm}	
		
		\noindent
		\hrulefill
		
    					\vspace{1mm}
		\noindent\textbf{Proof #1: }				
		
		\vspace{-2mm}
		\noindent
		\hrulefill
		
	}
	{$\hfill\square$
		
		\vspace{-2mm}
		\noindent
		\hrulefill
	\end{adjustwidth}
	\bigskip
}

\newtheorem{theorem}{Theorem}
\newtheorem{definition}{Definition}
\newtheorem{lemma}{Lemma}

\newtheorem{corollary}{Corollary}
\newtheorem{remark}{Remark}

\input{definitions.tex}

\usepackage[colorinlistoftodos, textwidth=4cm, shadow]{todonotes}

\newcommand{\hl}[1]{{\textit{#1}}}

\usepackage[colorinlistoftodos, textwidth=4cm, shadow]{todonotes}
\newcommand{\UCRL}{{\textcolor{red!50!black}{\texttt{UCRL2}}}}
\newcommand{\UCRLnew}{{\textcolor{red!50!black}{\texttt{UCRL3}}}}

\newcommand{\UCRLB}{{\textcolor{red!50!black}{\texttt{UCRL2B}}}}
\newcommand{\SCAL}{{\textcolor{red!50!black}{\texttt{SCAL}}}}
\newcommand{\PSRL}{{\textcolor{red!50!black}{\texttt{PSRL}}}}
\newcommand{\PSRLAVR}{{\textcolor{red!50!black}{\texttt{PSRL-AvR}}}}
\newcommand{\REGAL}{{\textcolor{red!50!black}{\texttt{REGAL}}}}
\newcommand{\KLUCRL}{{\textcolor{red!50!black}{\texttt{KL-UCRL}}}}

\newcommand{\EBF}{{\textcolor{red!50!black}{\texttt{EBF}}}}
\newcommand{\EVI}{\texttt{EVI}}

\renewcommand{\ln}{\log}

\usepackage{hyperref}       
\hypersetup{
  colorlinks  = true,    
  urlcolor     = blue,   
}

\usepackage[capitalize]{cleveref}

\icmltitlerunning{Tightening Exploration in Upper Confidence RL}

\begin{document}

\twocolumn[
\icmltitle{Tightening Exploration in Upper Confidence Reinforcement Learning}




\begin{icmlauthorlist}
\icmlauthor{Hippolyte Bourel}{inria}
\icmlauthor{Odalric-Ambrym Maillard}{inria}
\icmlauthor{Mohammad Sadegh Talebi}{diku}
\end{icmlauthorlist}

\icmlaffiliation{inria}{SequeL, Inria Lille -- Nord Europe, Villeneuve d'Ascq, France}
\icmlaffiliation{diku}{Department of Computer Science, University of Copenhagen, Copenhagen, Denmark}

\icmlcorrespondingauthor{Hippolyte Bourel}{hippolyte.bourel@ens-rennes.fr}
\icmlcorrespondingauthor{Odalric-Ambrym Maillard}{odalric.maillard@inria.fr}
\icmlcorrespondingauthor{Mohammad Sadegh Talebi}{sadegh.talebi@di.ku.dk}

\icmlkeywords{Machine Learning, ICML}

\vskip 0.3in
]



\printAffiliationsAndNotice{}  

\begin{abstract}
The upper confidence reinforcement learning (\UCRL) algorithm introduced in \citep{jaksch2010near} is a popular method to perform regret minimization in unknown discrete Markov Decision Processes under the average-reward criterion. 
Despite its nice and generic theoretical regret guarantees, 
this algorithm and its variants have remained until now mostly theoretical as numerical experiments in simple environments exhibit long burn-in phases before the learning takes place. 
In pursuit of practical efficiency, we present \UCRLnew, following the lines of \UCRL, but with two key modifications: First, it uses state-of-the-art time-uniform concentration inequalities to compute confidence sets on the reward and (component-wise) transition distributions for each state-action pair. Furthermore, to tighten exploration, it uses an adaptive computation of the support of each transition distribution, which in turn enables us to revisit the extended value iteration procedure of \UCRL\ to optimize over distributions with reduced support
	by disregarding low probability transitions, while still ensuring near-optimism.
	We demonstrate, through numerical experiments in standard environments, that reducing exploration this way yields a substantial numerical improvement compared to \UCRL\ and its variants.
	On the theoretical side, these key modifications enable us to derive a regret bound for \UCRLnew\ improving on \UCRL,
	that for the first time makes appear notions of local diameter and local effective support, thanks to variance-aware concentration bounds.	
\end{abstract}

\section{Introduction}\label{sec:intro}

In this paper, we consider  Reinforcement Learning (RL) in an unknown and discrete Markov Decision Process (MDP) under the average-reward criterion, when the learner interacts with the system
in a \emph{single, infinite} stream of observations, starting from an initial state without any reset.
More formally, let $M = (\cS,\cA,p,\nu)$ be an undiscounted MDP, where $\cS$ denotes the discrete state-space with cardinality $S$, and $\cA$ denotes the discrete action-space with cardinality $A$. $p$ is the transition kernel such that $p(s'|s,a)$ denotes the probability of transiting to state $s'$, starting from state $s$ and executing action $a$. We denote by $\cK_{s,a}$ the set of successor states of the state-action pair $(s,a)$, that is $\cK_{s,a}:= \{x\!\in\!\cS\!: p(x|s,a)\!>\!0\}$, and further define $K_{s,a}:= |\cK_{s,a}|$. Finally,  $\nu$ is a reward distribution function supported on $[0,1]$ with mean function denoted by $\mu$. The interaction between the learner and the environment  proceeds as follows. The learner starts in some state $s_1\in \cS$ at time $t = 1$. At each time step $t\in \mathbb N$, where the learner is in state $s_t$, she chooses an action $a_t\in \cA$ based on $s_t$ as well as her past decisions and observations. When executing action $a_t$ in state $s_t$, the learner receives a random reward $r_t:=r_t(s_t,a_t)$ drawn (conditionally) independently from distribution $\nu(s_t, a_t)$, and whose mean is $\mu(s_t,a_t)$. The state then transits to a next state $s_{t+1}\sim p(\cdot|s_t,a_t)$, and a new decision step begins.  For  background material on MDPs and RL, we refer to standard textbooks \citep{sutton1998reinforcement,puterman2014markov}.

The goal of the learner is to maximize the \textit{cumulative reward} gathered in the course of her interaction with the environment. The transition kernel $p$ and the reward function $\nu$ are initially \hl{unknown}, and so the learner has to learn them by trying different actions and recording the realized rewards and state transitions. The performance of the learner can be assessed through the notion of \emph{regret}, which compares the cumulative reward gathered by an oracle, being aware of $p$ and $\nu$, to that gathered by the learner. Following \citep{jaksch2010near}, we define the regret of a learning algorithm $\bA$ after $T$ steps as
$\kR(\bA,T) := Tg^\star - \sum_{t=1}^T r_t$, where $g^\star$ denotes the \emph{average-reward (or gain)} attained by an optimal policy. Alternatively, the objective of the learner is
to minimize the regret, which entails balancing exploration and exploitation.

To date, several algorithms have been proposed in order to minimize the regret based on the \emph{optimism in the face of uncertainty} principle, a.k.a.~the optimistic principle, originated from the seminal work \citep{lai1985asymptotically} on stochastic multi-armed bandits.
Algorithms designed based on this principle typically maintain confidence bounds on the unknown reward and transition distributions, and choose an optimistic model that leads to the highest average-reward.
A popular algorithm implementing the optimistic principle for the presented RL setup is \UCRL, which was introduced in the seminal work \citep{jaksch2010near}.  \UCRL\ achieves a non-asymptotic regret upper bound scaling as $\widetilde \cO(DS\sqrt{AT})$\footnote{The notation $\widetilde\Ocal(\cdot)$ hides terms that are poly-logarithmic in $T$.} with high probability, in any communicating MDP with $S$ states, $A$ actions, and diameter $D$.\footnote{Given an MDP $M$, the diameter $D:=D(M)$ is defined as $D(M) :=  \max_{s\neq s'}\min_\pi \EE[T^\pi(s,s')]$, where $T^\pi(s,s')$ denotes the number of steps it takes to get to $s'$ starting from $s$ and following policy $\pi$ \citep{jaksch2010near}.} \citet{jaksch2010near} also report a regret lower bound scaling as $\Omega(\sqrt{DSAT})$, indicating that the above regret bound for \UCRL\ is rate-optimal (up to logarithmic factors), i.e., it has a tight dependence on $T$, and can only be improved by a factor of, at most, $\sqrt{DS}$. 

Since the advent of \UCRL, several of its  variants have been presented in the literature; see, e.g., \citep{filippi2010optimism,bartlett2009regal,maillard2014hard,fruit2018efficient,talebi2018variance}. These variants mainly strive to improve the regret guarantee and/or empirical performance of \UCRL\ by using improved confidence bounds or planning procedures. Although these algorithms enjoy delicate and strong theoretical regret guarantees, their numerical assessments have shown that they typically achieve a bad performance even for state-spaces of moderate size. In particular, they suffer from a long burn-in phase before the learning takes place, rendering them impractical for state-spaces of moderate size.
It is natural to ask whether such a bad empirical performance is due to the main principle of \UCRL-style strategies,
 such as the optimistic principle, or to a not careful enough application of this principle. 
 For instance, in a different, episodic and Bayesian framework, \PSRL\ \citep{osband2013more} has been reported to significantly outperform \UCRL\ in numerical experiments. In this paper, we answer  this question by showing, perhaps surprisingly, that a simple but crucial modification of \UCRL\ that we call \UCRLnew\ significantly outperforms other variants, while preserving (an improving on) their theoretical guarantees. Though our results do not imply that optimistic strategies are the best, they show that they can be much stronger competitors than vanilla \UCRL.
\begin{figure*}[!tbh]
\begin{center}
\footnotesize
\vspace{-2mm}
\begin{tabular}{ccc}
	\text{Algorithm} & \text{Regret bound} & \text{Comment}\\
	\hline
	\UCRL\ \cite{jaksch2010near} & $\cO\Big(DS\sqrt{AT\log(T/\delta)}\Big)$ &  \\
	\KLUCRL\ \cite{filippi2010optimism} & $\cO\Big(DS\sqrt{AT\log(\log(T)/\delta)}\Big)$ &  Valid for fixed $T$ provided as input. \\
    \KLUCRL\ \cite{talebi2018variance} & $\cO\Big(\Big[D\!+\!\sqrt{S\sum_{s,a} (\Var_{s,a}\!\lor\!1)}\Big]\sqrt{T\log(\log(T)/\delta)}\Big)$ & Restricted to ergodic MDPs.  \\    
	\SCAL$^+$\ \text{\cite{qian2019exploration}} & $\cO\Big(D\sqrt{\sum_{s,a} K_{s,a} T\log(T/\delta)}\Big)$ & Without knowledge of the span. \\
	\UCRLB\ \text{\cite{fruit2019improved}}& $\cO\Big(\sqrt{D\sum_{s,a} K_{s,a} T\log(T)\log(T/\delta)}\Big)$ &  Note the extra $\sqrt{\log(T)}$ term.\\
	\UCRLnew\ \textbf{(This Paper)} & $\cO\Big(\big(D\!+\!\sqrt{\sum_{s,a} (D_s^2L_{s,a}\!\lor\!1)}\big)\sqrt{T\log(T/\delta)}\Big)$ & \\
    Lower Bound \cite{jaksch2010near} & $\Omega\big(\sqrt{DSAT}\big)$ & \\
\hline
\end{tabular}
\vspace{-2mm}
 \caption{Regret bounds of state-of-the-art algorithms for average-reward reinforcement learning. Here, $x\lor y$ denotes the maximum between $x$ and $y$. For \KLUCRL, $\Var_{s,a}$ denotes the variance of the optimal bias function of the true MDP, when the state is distributed according to $p(\cdot|s,a)$. For \UCRLnew, $L_{s,a}:= \big(\sum_{x\in \cS} \sqrt{p(x|s,a)(1 - p(x|s,a))}\big)^2$ denotes the local effective support of $p(\cdot|s,a)$.} \vspace{-3mm}
 \label{table:alg_summary}
\end{center}
\end{figure*}

\paragraph{Contributions.} We introduce \UCRLnew, a refined variant of \UCRL, whose design combines the following key elements: First, it uses tighter confidence bounds on components of the transition kernel (similarly to \citep{dann2017unifying}) that are \emph{uniform in time}, a property of independent interest for algorithm design in other RL setups; we refer to Section~\ref{sec:confidence_bounds} for a detailed presentation.
More specifically, for each component of a next-state transition distribution, it uses one time-uniform concentration inequality for $[0,1]$-bounded observations and one for Bernoulli distributions with a Bernstein flavor.


The second key design of the algorithm is a novel procedure, which we call \texttt{NOSS}\footnote{Near-Optimistic Support Optimization}, that adaptively computes an estimate of the support of transition probabilities of various state-action pairs. Such estimates are in turn  used to compute a near-optimistic value and policy (Section~\ref{sub:noss}). Combining \texttt{NOSS} with the Extended Value Iteration (\EVI) procedure, used for planning in \UCRL, allows us to devise \texttt{EVI-NOSS}, which is a refined variant of \EVI. 
This step is non-trivial as it requires to find a near-optimistic, as opposed to \emph{fully optimistic}, policy. 
 Furthermore, this enables us to make appear in the regret analysis notions of \emph{local diameter} (Definition \ref{def:local_diameter}) as well as  \emph{local effective support} (Section \ref{sec:regret_UCRL3}), which in turn leads to a more problem-dependent regret bound. We define the local diameter below.
 
\begin{definition}[Local diameter of state $s$]
\label{def:local_diameter}
	Consider state $s\in \cS$. For $s_1,s_2\in \cup_{a\in \cA} \cK_{s,a}$ with $s_1\neq s_2$, let $T^\pi(s_1,s_2)$ denote the number of steps it takes to get to $s_2$ starting from $s_1$ and following policy $\pi$. Then, the local diameter of MDP $M$ for $s$, denoted by $D_{s}:=D_{s}(M)$,  is defined as
	$$
	D_{s}:= \max_{s_1,s_2 \in  \cup_a \cK_{s,a}}   \min_\pi \Esp[T^\pi(s_1,s_2)].
	$$
\end{definition}

\vspace{-2mm}\noindent
On the theoretical side, we show in Theorem~\ref{thm:regret_UCRLnew} that \UCRLnew\ enjoys a regret bound scaling similarly to that established for the best variant of \UCRL\ in the literature as in, e.g., \citep{fruit2018efficient}. For better comparison with other works, we make sure to have an explicit bound including small constants for the leading terms. 
Thanks to a refined and careful analysis that we detail in the appendix,  we also improve on the lower-order terms of the regret that we show should not be overlooked in practice.
We provide in Section~\ref{sec:NumExp} a detailed comparison of the leading terms involved in several state-of-the-art algorithms to help better understand the behavior of these bounds.
We also demonstrate through numerical experiments on standard environments that combining these 
 refined, state-of-the-art confidence intervals together with \texttt{EVI-NOSS} 
 yield a substantial improvement over \UCRL\ and its variants.
 In particular, \UCRLnew\ admits a burn-in phase, which is smaller than that of \UCRL\ by an order of magnitude. 

\paragraph{Related work.}
The study of RL under the average-reward criterion dates back to the seminal papers \citep{graves1997asymptotically} and \citep{burnetas1997optimal}, followed by \citep{tewari2008optimistic}. Among these studies, for the case of ergodic MDPs, \citet{burnetas1997optimal} derive an asymptotic MDP-dependent lower bound on the regret, and provides an asymptotically optimal algorithm.
Algorithms with finite-time regret guarantees and for wider classes of MDPs are presented in \citep{auer2007logarithmic,jaksch2010near,bartlett2009regal,filippi2010optimism,maillard2014hard,talebi2018variance,fruit2018near,fruit2018efficient,zhang2019regret,qian2019exploration}.
Among these works, \citet{filippi2010optimism} introduce \KLUCRL, which is a variant of \UCRL\ that uses the KL divergence to define confidence bounds. Similarly to \UCRL, \KLUCRL\ achieves a regret of $\widetilde\Ocal(DS\sqrt{AT})$ in communicating MDPs. A more refined regret bound for \KLUCRL\ in ergodic MDPs is presented in \citep{talebi2018variance}. \citet{bartlett2009regal} present \REGAL\ and report a $\widetilde\Ocal(D' S\sqrt{AT})$ regret with high probability in the larger class of weakly communicating MDPs, provided that the learner knows  an upper bound  $D'$ on the span of the optimal bias function of the true MDP. \citet{fruit2018efficient} present \SCAL, which similarly to \REGAL\ works in weakly communicating MDPs, but  admits an efficient implementation. A similar algorithm called \SCAL$^+$ is presented in \cite{qian2019exploration}. Both \SCAL\ and \SCAL$^+$ admit a regret bound scaling as $\widetilde \cO\Big(D\sqrt{\sum_{s,a} K_{s,a}T}\Big)$. In a recent work, \citet{zhang2019regret} present \EBF\ achieving a regret of $\widetilde\cO\big(\sqrt{HSAT}\big)$ assuming that the learner knows an upper bound $H$ on the span of the optimal bias function of the true MDP.\footnote{We remark that the universal constants of the leading term here are fairly large.} However, \EBF\ does not admit a computationally efficient implementation.
 
Another related line of works considers posterior sampling methods such as  \cite{osband2013more} inspired by Thompson sampling \cite{thompson1933likelihood}. For average-reward RL, existing works on these methods report Bayesian regret bounds, with the exception  of \cite{agrawal2017optimistic}, whose corrected regret bound, reported in \citep{agrawal2017posterior}, scales as $O(DS\sqrt{AT} \log^3(T))$ and is valid for $T\ge S^4A^3$.

We finally mention that some studies consider regret minimization in MDPs in the \textit{episodic} setting, with a fixed and known horizon; see, e.g., \citep{osband2013more,azar2017minimax,dann2017unifying,efroni2019tight,zanette2019tighter}. Despite some similarity between the episodic and average-reward settings, the techniques developed for the episodic setting in these papers strongly rely on the fixed length of the episode. Hence, the tools in these papers do not directly carry over to the case of average-reward RL considered here 
(in particular, when closing the gap between lower and upper bounds is concerned).

In Figure \ref{table:alg_summary}, we report regret upper bounds of state-of-the-art algorithms for average-reward RL. We do not report \REGAL\ and \EBF\ in this table, as no corresponding efficient implementation is currently known.  Furthermore, we stress that  the presented regret bound for \UCRLnew\ does not contradict the worst-case lower bound of $\Omega(\sqrt{DSAT})$ presented in \citep{jaksch2010near}. Indeed, for the worst-case MDP used to establish this lower bound in \citep{jaksch2010near}, both the local and global diameters coincide.

\paragraph{Notations.} We introduce some notations that will be used throughout. For $x,y\in \RR$, $x\lor y$ denotes the maximum between $x$ and $y$. $\Delta_\cS$ represents the set of all probability distributions defined on $\cS$. For a distribution $p\in\Delta_\cS$ and a vector-function $f = (f(s))_{s\in\cS}$,  we let $Pf$ denote its application on $f$, defined by $Pf=\Esp_{X\sim p}[f(X)]$. We introduce 
$\Delta_{\cS}^{\cS\times\cA}:=\{q: q(\cdot|s,a) \in \Delta_{\cS}, (s,a)\in \cS\times \cA\}$, and for $p\in\Delta_\cS^{\cS\times\cA}$, we define the corresponding operator $P$ such that $Pf: s,a\mapsto  \Esp_{X\sim p(\cdot|s,a)}[f(X)]$.
We also introduce $\bS(f) = \max_{s\in \cS} f(s) - \min_{s\in \cS} f(s)$. 

Under a given algorithm and for a pair $(s,a)$, we denote by $N_t(s,a)$ the total number of observations of $(s,a)$ up to time $t$, and if $(s,a)$ is not sampled yet by $t$, we set $N_t(s,a)=1$. Namely,  $N_t(s,a):=1 \lor\sum_{t'=1}^{t-1} \indic{(s_{t'},a_{t'})=(s,a)}$. Let us define $\widehat \mu_{t}(s,a)$ as the empirical mean reward built using $N_t(s,a)$ i.i.d.~samples from $\nu(s,a)$  (and whose mean is $\mu(s,a)$), and $\widehat p_t(\cdot|s,a)$ as the empirical distribution built using $N_t(s,a)$ i.i.d.~observations from $p(\cdot|s,a)$. 

\section{Background: The \UCRL\ Algorithm}
Before presenting \UCRLnew\ in Section \ref{sec:UCRL3}, we briefly present \UCRL\ \citep{jaksch2010near}. To this end, let us introduce the following two sets: For each $(s,a)\in \cS\times \cA$, 
\als{
&c^{\text{\UCRL}}_{t,\delta}(s,a) =\\
&\, \bigg\{\mu'\in[0,1]:|\widehat \mu_t(s,a)  - \mu'| \le \sqrt{\frac{3.5\log(\tfrac{2SAt}{\delta})}{N_t(s,a)}} \bigg\}\,,\\
&\cC^{\text{\UCRL}}_{t,\delta}(s,a) =\\
&\, \bigg\{ p'\in \Delta_{\cS}: \|\widehat p_{t}(\cdot|s,a)  - p' \|_1 \leq \sqrt{\frac{14S\log(\tfrac{2At}{\delta})}{N_t(s,a)}}  \bigg\}\,.
}
At a high level, \UCRL\ maintains the set of MDPs $\cM_{t,\delta}=\{\widetilde M=(\cS,\cA, \widetilde p, \widetilde \nu)\}$, where for each $(s,a)\in\cS\times\cA$, $\widetilde p(\cdot|s,a) \in \cC^{\text{\UCRL}}_{t,\delta}(s,a)$ and $\widetilde \mu(s,a)\in c^{\text{\UCRL}}_{t,\delta}(s,a)$ (with $\widetilde \mu$ denoting the mean of $\widetilde\nu$). 
It then implements the optimistic principle by trying to compute
$\overline{\pi}_t^+ = \argmax_{\pi:\cS\to\cA} \max\{ g_\pi^M:  M \in \cM_{t,\delta} \}$, where $g_\pi^M$ is the average-reward (or gain) of policy $\pi$ in MDP $M$. This is carried out approximately by \EVI\ that builds a near-optimal policy $\pi^+_t$ and an MDP $\widetilde M_t$ such that $g_{\pi^+_t}^{\widetilde M_t}  \geq \max_{\pi,  M\in\cM_{t,\delta} }g_\pi^M - \tfrac{1}{\sqrt{t}}$. Finally,  \UCRL\ does not recompute $\pi^+_t$ at each time step. Instead, it proceeds in internal episodes,  indexed by $k\in \NN$, 
 where a near-optimistic policy $\pi^+_{t}$ is computed only at the starting 
time of each episode. Letting $t_k$ denote the starting time of episode $k$, the algorithm computes
$\pi_k^+ :=\pi^+_{t_k}$ and applies it until $t = t_{k+1} - 1$, where 
the sequence $(t_k)_{k\geq 1}$ is defined as follows: $t_1 = 1$, and for all $k > 1$, 
$$
t_k  = \min\bigg\{ \!t \!> \!t_{k-1}:  \max_{s,a} \frac{v_{t_{k-1}:t}(s,a)}{N_{t_{k-1}}(s,a)}\!\geq\!1 \bigg\},
$$
where $v_{t_1:t_2}(s,a)$ denotes the number of observations of pair $(s,a)$ between time $t_1$ and $t_2$. The \EVI\ algorithm writes as presented in Algorithm~\ref{alg:EVI}.

\begin{algorithm}[!hbtp]	
	\caption{Extended Value Iteration (\EVI)}
	\label{alg:EVI}
    \footnotesize
\begin{algorithmic}
	\REQUIRE $\epsilon_t$
		\STATE Let $u_0\equiv 0, u_{-1}\equiv-\infty$, $n=0$
		\WHILE{$\bS(u_{n}-u_{n-1}) > \epsilon_t$}
		\STATE $\!\!\!$Compute
		$\begin{cases}
	 \mu^+: s,a\mapsto\max \{\mu':  \mu' \!\in\! c^\UCRL_{t,\delta}(s,a)\}\\
		p^+_{n}: s,a\mapsto \argmax \{ P'u_n:  p'\!\in\! \cC^\UCRL_{t,\delta}(s,a)\}		
		\end{cases}$
		\STATE $\!\!\!$Update
	$\!\begin{cases}
\!u_{n+1}\!(s) =   \max\{ \mu^+\!(s,a) \!+\! (P^+_{n} u_n)(s,a)\!\!: {a\!\in\!\!\cA}\}\\
\!\pi^+_{n+1}\!(s) \!\in\! \Argmax\{\mu^+\!(s,a) \!+\! (P^+_{n} u_n)(s,a)\!\!:{a\!\in\!\!\cA}\}
		\end{cases}$
		\STATE $\!\!\!n=n+1$
\ENDWHILE
	\end{algorithmic}
\normalsize
\end{algorithm}

\vspace{-3mm}
\section{The \UCRLnew\ Algorithm} \label{sec:UCRL3}
In this section, we introduce the \UCRLnew\ algorithm, a variant of \UCRL\ that relies on two main ideas motivated as follows:

(i) While being a theoretically appealing strategy, \UCRL\ suffers from conservative confidence intervals, yielding an unacceptable empirical performance. Indeed, in the design of \UCRL, the random stopping times $N_t(s,a)$  are handled using simple union bounds, resulting in loose confidence bounds. The first modification we introduce has thus the same design as \UCRL, but replaces these confidence bounds with those derived from tighter time-uniform concentration inequalities. Furthermore, unlike \UCRL, \UCRLnew\ does not use the $L_1$ norm to define the confidence bound of transition probabilities $p$. Rather it defines confidence bounds for each transition probability $p(s'|s,a)$, for each pair $(s,a)$, similarly to \SCAL\ or \UCRLB. Indeed, one drawback of $L_1$-type confidence bounds is that they require an upper bound on the size of the support of the distribution. Without further knowledge, only the conservative bound of $S$ on the support can be applied. In \UCRL, this causes a factor $S$ to appear inside the square-root, due to a union bound over $2^S$ terms. Deriving $L_1$-type confidence bounds adaptive to the support size seems challenging. In stark contrast, entry-wise confidence bounds can be used without knowing the support: when $p(\cdot|s,a)$ has a support much smaller than $S$, this may lead to a substantial improvement. Hence,  \UCRLnew\ relies on time-uniform Bernoulli concentration bounds (presented in Section \ref{sec:confidence_bounds} below).

(ii)
In order to further tighten exploration, the second idea behind \UCRLnew\ is to revisit \EVI\ to compute a near-optimistic policy. Indeed, the optimization procedure used in \EVI\ considers all plausible transition probabilities without support restriction, causing unwanted exploration. We introduce  a novel value iteration procedure, called \texttt{EVI-NOSS}, which uses a restricted support optimization, where the considered support is chosen adaptively in order to retain near-optimistic guarantees.

We discuss these two modifications below in greater detail.

\subsection{Confidence Bounds}\label{sec:confidence_bounds}
We introduce the following high probability confidence sets for the mean rewards: For each $(s,a)\in \cS\times \cA$,
\als{
c_{t,\delta_0}(s,a)\!=\!\Big\{\mu'\in[0,1]:|\widehat \mu_t(s,a)  - \mu'| \le b^r_{t,\delta_0/(SA)}(s,a)\Big\}\,,
}
where  we introduced the notation
\als{
b^r_{t,\delta_0/(SA)}&(s,a):= \max\bigg\{\tfrac{1}{2} \beta_{N_t(s,a)}\big(\tfrac{\delta_0}{SA}\big), \\
&\sqrt{\frac{2\widehat\sigma_t^2(s,a)}{N_t(s,a)}\ell_{N_t(s,a)}\big(\tfrac{\delta_0}{SA}\big)} \!+\! \frac{7\ell_{N_t(s,a)}\big(\tfrac{\delta_0}{SA}\big)}{3N_t(s,a)}\bigg\}\,,
}
with $\widehat\sigma^2_t(s,a)$ denoting the empirical variance of the reward function of $(s,a)$ built using the observations gathered up to time $t$, and 
where $\ell_n(\delta) \!=\! \eta\log\Big(\frac{\log(n)\log(\eta n)}{\log^2(\eta)\delta}\Big)$ with $\eta\!=\!1.12$,\footnote{Any $\eta>1$ is valid, and $\eta=1.12$ yields a small bound.}
and  
$\beta_n(\delta):=\sqrt{ \frac{2(1+\frac{1}{n})\log(\sqrt{n+1}/\delta)}{n}}$. 

The definition of this confidence set is motivated by Hoeffding-type concentration inequalities for $1/2$-sub-Gaussian distributions\footnote{We recall that random variables bounded in $[0,1]$ are $\tfrac{1}{2}$-sub-Gaussian.}, modified to hold for an arbitrary random stopping time, using the method of mixtures (a.k.a.~the Laplace method) from \citep{pena2008self}.
This satisfies by construction that
$$
\Pr\Big(\exists t\in\Nat,(s,a)\in\cS\times\cA,\,\, \mu(s,a)\notin c_{t,\delta_0}(s,a)\Big)\leq 3\delta_0.
$$
We recall the proof of this powerful result for completeness in Appendix~A.
Regarding the transition probabilities, we introduce the two following sets: For each $(s,a,s')\in \cS\times \cA\times \cS$,

\vspace{-8mm}
\als{
&C_{t,\delta_0}(s,a,s')\!=\! \bigg\{\! q\!\in\![0,1]\!:  \\
&
| \widehat p_{t}(s'|s,a) \! -\! q |
 \!\le\! \sqrt{\frac{2q(1\!-\!q)}{N_t(s,a)}\ell_{N_t(s,a)}\big(\tfrac{\delta_0}{SA}\big)} \!+\! \frac{\ell_{N_t(s,a)}\big(\tfrac{\delta_0}{SA}\big)}{3N_t(s,a)}, \\
&\qquad \qquad\text{ and } -\sqrt{\underline{g}(q)}\!\le  \frac{\widehat p_{t}(s'|s,a)  \!-\! q}{\beta_{N_t(s,a)}\big(\tfrac{\delta_0}{SA}\big)} \!\le\! \sqrt{g(q)}
\bigg\}\,,
}
where $\underline{g}(p)\!=\!\begin{cases} g(p)&\text{if }p\!<\!0.5\\ p(1\!-\!p) &\text{else}\end{cases}\!,$ with $g(p) \!=\! \frac{1/2-p}{\log(1/p\!-\!1)}$. 
The first inequality  comes from the Bernstein concentration inequality, modified using a peeling technique in order to handle arbitrary random stopping times. We refer the interested reader to \citep{maillard2019mathematics} for the generic proof technique behind this result. 
\citet{dann2017unifying} use similar proof techniques for Bernstein's concentration, however the resulting bounds are looser; we discuss this more in Appendix~A.3.
The last two inequalities are obtained by applying again the method of mixture for sub-Gaussian random variables, with a modification: Indeed, Bernoulli random variables are not only $1/2$-sub-Gaussian, but satisfy a stronger sub-Gaussian tail property, already observed in
\citep{berend2013concentration,raginsky2013concentration}. 
We discuss this in great detail in Appendix~A.2.

\UCRLnew\ finally considers the set of plausible MDPs $\cM_{t,\delta}=\{ \widetilde M= (\cS,\cA,\widetilde p,\widetilde \nu)\}$, where
for each $(s,a)\in\cS\times\cA$,
\al{
\label{eq:UCRLnew_CB}
&\widetilde \mu(s,a) \in c_{t,\delta_0}(s,a),\\
&\widetilde p(\cdot|s,a) \!\in\!\cC_{t,\delta_0}(s,a)\!=\!\bigg\{\!p'\!\in\!\Delta_\cS\!\!: \forall s'\!, p'(s') \!\in\! C_{t,\delta_0}(s,a,s'\!) \! \bigg\}.\nonumber
}
Finally, the confidence level is chosen as\footnote{When an upper bound $\overline{K}$ on $\max_{s,a} K_{s,a}$ is known, one could choose the confidence level $\delta_0 = \delta/(3+3\overline{K})$.} $\delta_0 = \delta/(3+3S)$. 

\begin{lemma}[Time-uniform confidence bounds]\label{lem:CI_has_trueMDP}
	For any MDP with rewards bounded in $[0,1]$, mean reward function $\mu$, and transition function $p$,
	for all $\delta\in(0,1)$, it holds
\als{
&\Pr\bigg(\exists t\in\Nat, (s,a)\in\cS\times\cA, \\
&\quad  \mu(s,a) \notin c_{t,\delta_0}(s,a)
	\,\,\text{ or } \,\,p(\cdot|s,a) \notin \cC_{t,\delta_0}(s,a) \bigg) \leq \delta\,.
}
\end{lemma}

\subsection{Near-Optimistic Support-Adaptive Optimization}\label{sub:noss}
Last, we revisit the \EVI\ procedure of \UCRL.
When computing an optimistic MDP, \EVI\ uses for each pair $(s,a)$ an optimization over the set of all plausible transition probabilities (that is, over all distributions $q\in\cC_{t,\delta}(s,a)$).
This procedure comes with no restriction on the support of the considered distributions.
In the case where $p(\cdot|s,a)$ is supported on a sparse subset of $\cS$, this may however lead to computing an optimistic distribution with a large support, which in turn results in unnecessary exploration, and thereby degrades the performance. 
The motivation to revisit \EVI\ is to provide a more adaptive way of handling sparse supports.

Let $\widetilde \cS\subset\cS$ and $f$ be a given function (intuitively, the value function $u_i$ at the current iterate $i$ of \EVI), and consider the  following optimization problem for a specific state-action pair $(s,a)$:
\beqa
\overline{f}_{s,a}(\widetilde \cS)=
\max_{\widetilde p \in \cX} \sum_{s'\in\widetilde \cS} f(s')\widetilde p(s')\, ,\quad \text{where }\label{eqn:fbar}
\eeqa

\vspace{-7mm}
\als{
\cX \!=\! \bigg\{\widetilde p\!: \forall s'\!\in\!\widetilde \cS,\,\widetilde{p}(s') \!\in\! C_{t,\delta}(s,a,s') \text{ and }\! \sum_{s'\in\widetilde\cS}\widetilde p(s') \!\leq\! 1\bigg\}\,.
}

\begin{remark}[Optimistic value]\label{rem:nearoptsupport}
	The quantity $\overline{f}_{s,a}(\widetilde \cS)$ is conveniently defined by an optimization over positive measures whose mass may be less than one. The reason is that $p(\widetilde \cS|s,a)\leq 1$ in general.
	This ensures that $p(\cdot|s,a)\!\in\!\cX$  indeed holds with high probability, and thus $\overline{f}_{s,a}(\widetilde \cS)\geq \sum_{s'\in\widetilde \cS}f(s')p(s'|s,a)$ as well.
\end{remark}
%

The original \EVI\  procedure (Algorithm \ref{alg:EVI}) computes  $\overline{f}_{s,a}(\cS)$ for the function $f=u_i$ at each iteration $i$.
When $p=p(\cdot|s,a)$ has a sparse support included in $\widetilde \cS$, $C_{t,\delta}(s,a,s')$ often does not reduce to $\{0\}$ for $s'\notin\widetilde \cS$, while one may prefer to force a solution with a sparse support.
A naive way to proceed is to define $\widetilde \cS$ as the empirical support (i.e., the support of $\widehat p_t(\cdot|s,a)$).
Doing so, one however solves a \textit{different} optimization problem than the one using the full set $\cS$, which means we may lose the optimistic property (i.e., $\overline{f}_{s,a}(\widetilde \cS) \geq \Esp_{X\sim p(\cdot|s,a)}[f(X)]$ may not hold) and get an uncontrolled error.
Indeed, the following decomposition
\beqan
\Esp_{X\sim p}[f(X)]
&=& \sum_{s'\in\widetilde \cS} f(s')p(s') +
\underbrace{\sum_{s'\notin\widetilde \cS} f(s')p(s')}_{\text{error}}\,,
\eeqan

\vspace{-3mm}\noindent
shows that computing an optimistic value restricted on $\widetilde \cS$ only upper bounds the first term in the right-hand side. The second term (the error term) needs to be upper bounded as well.
Consider a pair $(s,a)$, $t\ge 1$, and let $n:=N_t(s,a)$. Provided that $\widetilde S$ contains the support of $\widehat p_t$, thanks to Bernstein's confidence bounds, it is easy to see\footnote{They are of the form $p'- \widehat p_n(s') \leq a\sqrt{p'} + b$ where $a=\widetilde \Theta(n^{-1/2})$
	and $b = \widetilde\Theta(n^{-1})$.
This implies that for $s'$ outside of  the support of $\widehat p_n$,  $p'\leq a\sqrt{p'} + b$, that is $p' \leq (\sqrt{a/4}+\sqrt{a/4+b})^2$.}
 that the first term  in the above decomposition contains terms scaling as $\widetilde \cO(n^{-1/2})$, 
 while the error term contains only terms scaling as $\widetilde \cO(n^{-1})$. 
On the other hand, the error term  sums $|\cS\!\setminus\!\widetilde\cS|$ many elements, which can be large in case $p$ is sparse, and thus may even exceed $\overline{f}_{s,a}(\widetilde \cS)$ for small $n$. To ensure the error term does not dominate the first term, we introduce the Near-Optimistic Support-adaptive Optimization (\texttt{NOSS}) procedure, whose generic pseudo-code is presented  in Algorithm~\ref{alg:SupportSelection}. For instance, for a given pair $(s,a)$ and time $t$, \texttt{NOSS} takes as input a target function $f=u_i$ (i.e., the value function at iterate $i$), the support $\widehat \cS$ of the empirical distribution $\widehat p_t(\cdot|s,a)$, high-probability confidence sets $\cC:=\{C_{t,\delta}(s,a,s'), s'\in \cS\}$, and a parameter $\kappa\in (0,1)$. It then adaptively augments $\widehat \cS$ in order to find a set $\widetilde S$, whose corresponding value function $\overline{f}_{s,a}(\widetilde \cS)$ is near-optimistic, as formalized in the following lemma:

\begin{algorithm}[!hbtp]
	\caption{\texttt{NOSS}$(f,\widehat \cS,\cC,\kappa)$} \label{alg:SupportSelection}
	\footnotesize
    \begin{algorithmic}
		\STATE Let $\widetilde \cS = \widehat \cS \cup \argmax_{s\in\cS} f(s)$, and define $\overline{f}$ using $f$ and confidence sets $\cC$ (see \eqref{eqn:fbar}).
		\WHILE{ $\overline{f}(\cS\setminus\widetilde\cS)\geq \min(\kappa,\overline{f}(\widetilde \cS))$ }
		\STATE Let $\tilde s\in \Argmax_{s\notin\widetilde \cS} f(s)$
		\STATE $\widetilde \cS = \widetilde \cS \cup \{\tilde s\}$
		\ENDWHILE
		\STATE \textbf{return} $\widetilde \cS$
	\end{algorithmic}
\end{algorithm}

\begin{algorithm}[!ht]
	\caption{\texttt{EVI-NOSS}$(p, c, \cC, N_{\max},\epsilon)$}
	\label{alg:EVI-NOSS}
    \footnotesize	
\begin{algorithmic}
		\STATE Let $u_0\equiv 0, u_{-1}\equiv-\infty$, $n=0$
		\WHILE{$\bS(u_{n}-u_{n-1}) > \epsilon$}
		\STATE Compute for all $(s,a)$:  
		\STATE $\widetilde \cS_{s,a} = \texttt{NOSS}(u_n - \min_s u_n, \supp(p(\cdot|s,a)), \cC, \kappa)$, with $\kappa= 10\bS(u_n)|\supp(p(\cdot|s,a))|/N_{\max}^{3/2}$ 
		\STATE $\widetilde\cC(s,a) = \big\{p'\in \cC(s,a):  p'(x)=0,\forall x\in \cS\!\setminus\!\widetilde\cS_{s,a}\big\}$ 		
		\STATE $\!\!\!$Compute
	$\!\begin{cases}
\!\mu^+\!: s,a\mapsto\max \{\mu':  \mu' \!\in\! c(s,a)\}\\
p^+_{n}: s,a\mapsto \argmax \{ P'u_n:  p'\!\in\! \widetilde\cC(s,a)\}
		\end{cases}$
				\STATE $\!\!\!$Update
	$\!\begin{cases}
\!u_{n+1}\!(s) =   \max\{ \mu^+\!(s,a) \!+\! (P^+_{n} u_n)(s,a)\!\!: {a\!\in\!\!\cA}\}\\
\!\pi^+_{n+1}\!(s) \!\in\! \Argmax\{\mu^+\!(s,a) \!+\! (P^+_{n} u_n)(s,a)\!\!:{a\!\in\!\!\cA}\}
		\end{cases}$		
		\STATE $\!\!\!n=n+1$
\ENDWHILE
	\end{algorithmic}
\normalsize
\end{algorithm}


\begin{lemma}[Near-optimistic support selection]
\label{lem:near_opt_supp_selection}
	Let $\widetilde S$ be a set output by \texttt{NOSS}. Then, with probability higher than $1-\delta$,
	\beqan
 \overline{f}_{s,a}(\widetilde \cS)\!\geq\!	\Esp_{X\sim p(\cdot|s,a)}[f(X)] - \min\!\big\{\!\kappa,\overline{f}_{s,a}(\widetilde \cS), \overline{f}_{s,a}(\cS\!\setminus\!\widetilde \cS)\big\}\!.
	\eeqan
	In other words, the value function $\overline{f}_{s,a}(\widetilde \cS)$ is near-optimistic.
\end{lemma}

\paragraph{Near-optimistic value iteration: The \texttt{EVI-NOSS} algorithm.}
In \UCRLnew, we thus naturally revisit the \EVI\ procedure
and combine the following step at each iterate $n$ of \EVI\
\beqan
		p^+_{n}: s,a\mapsto &\argmax \{ P'u_n,  p'\in \cC_{t,\delta}(s,a)\}\,,
\eeqan
with \texttt{NOSS}: For a state-action pair $(s,a)$, \UCRLnew\  applies \texttt{NOSS} (Algorithm~\ref{alg:SupportSelection})
to the function $u_n-\min_s u_n(s)$ (i.e., the relative optimistic value function)
and empirical distribution $\widehat p_{t}(\cdot|s,a)$. We refer to the resulting algorithm as \texttt{EVI-NOSS}, as it combines \EVI\ with \texttt{NOSS}, and present its pseudo-code in Algorithm \ref{alg:EVI-NOSS}. Finally, for iterate $n$ in \texttt{EVI-NOSS}, we set the value of $\kappa$ to
\al{
\label{eq:kappa}
 \kappa = \kappa_{t,n}(s,a) = \frac{\gamma\bS(u_n)|\supp(\widehat p_t(\cdot|s,a))|}{{\max_{s,a} N_t(s,a)^{2/3}}},\text{ where } \gamma=10.
}

\vspace{-3mm}\noindent
	The scaling with the size of  support and the span of the considered function is intuitive. The reason to further normalize by
	${\max_{s',a'}N_t(s',a')^{2/3}}$ is to deal with the case when $N_t(s,a)$ is small:
	First, in the case of Bernstein's bounds, and since $\widetilde \cS$ contains at least the empirical support,
	$\min\big\{\overline{f}_{s,a}(\widetilde \cS), \overline{f}_{s,a}(\cS\setminus\widetilde \cS)\big\}$ should essentially scale as $\widetilde \cO(N_t(s,a)^{-1})$.
Hence for pairs such that $N_t(s,a)$ is large, $\kappa$ is redundant. Now for pairs that are not sampled a lot, $N_t(s,a)^{-1}$ may still be large even for large $t$, resulting in a possibly uncontrolled error. Forcing a $\max_{s,a}N_t(s,a)^{2/3}$ scaling ensures the near-optimality of the solution is preserved with enough accuracy to keep the cumulative regret controlled.
	This is summarized in the following lemma, whose proof is deferred to Appendix~B.

\begin{lemma}[Near-optimistic value iteration]\label{lem:noEVI}
	Using the stopping criterion $\bS(u_{n+1}-u_n) \leq \epsilon$, the \texttt{EVI-NOSS} algorithm satisfies that the average-reward (gain) $g^+_{n+1}$ of the policy $\pi_{n+1}^+$ and the MDP $\widetilde M = (\cS,\cA,\mu_{n+1}^+, p_{n+1}^+)$ computed at the last iteration $n+1$ is near-optimistic, in the sense that with probability higher than $1-\delta$, uniformly over all $t$, $g^+_{n+1} \geq g^\star - \epsilon - \overline\kappa$, where $\overline\kappa = \overline\kappa_{t,n} = \frac{\gamma\bS(u_n)K}{{\max_{s,a} N_t(s,a)^{2/3}}}$.
\end{lemma}

\begin{algorithm}[!hbtp]
   \caption{\UCRLnew\ with input parameter $\delta\in (0,1)$}
   \label{alg:ucrl3}
   \footnotesize
\begin{algorithmic}
   \STATE \textbf{Initialize:} For all $(s,a)$, set $N_0(s,a)=0$ and $v_0(s,a)=0$. Set $\delta_0=\delta/(3+3S)$. Set $t_0=0$, $t=1$, $k=1$.
   \FOR{episodes $k=1,2,\ldots$}
       \STATE Set $t_k = t$
       \STATE Set $N_{t_k}(s,a) = N_{t_{k-1}}(s,a)+ v_{k-1}(s,a)$ for all $(s,a)$
       \STATE Compute empirical estimates $\widehat \mu_{t_k}(s,a)$ and $\widehat p_{t_k}(\cdot|s,a)$ for all $(s,a)$
       \STATE Using Algorithm \ref{alg:EVI-NOSS}, compute 
       $$\pi^+_{t_k}\!=\!\texttt{EVI-NOSS}\Big(\widehat p_{t_k}, c_{t_k,\delta_0}, \cC_{t_k,\delta_0}, \max_{s,a} N_{t_k}(s,a),\tfrac{1}{\sqrt{t_k}}\Big)$$ 
		\STATE Set $v_k(s,a)=0$ for all $(s,a)$       
       \WHILE{$v_{k}(s_t,\pi_{t_k}^+(s_t))<  N_{t_k}(s_t,\pi_{t_k}^+(s_t))$}
            \STATE Observe the current state $s_t$, play action $a_t=\pi_{t_k}^+(s_t)$, and receive reward $r_t$
            \STATE Set $v_k(s_t,a_t)=v_k(s_t,a_t)+1$
            \STATE Set $t=t+1$
       \ENDWHILE
   \ENDFOR
\end{algorithmic}
\normalsize
\end{algorithm}

The pseudo-code of \UCRLnew\ is provided in Algorithm \ref{alg:ucrl3}.

\subsection{Regret Bound of \UCRLnew}
\label{sec:regret_UCRL3}
We are now ready to present a finite-time regret bound for \UCRLnew. 
Before presenting the regret bound in Theorem \ref{thm:regret_UCRLnew} below, we introduce the notion of \emph{local effective support}. 
Given a pair $(s,a)$, we define the \emph{local effective support} $L_{s,a}$ of $(s,a)$ as: 
$$ L_{s,a} := \Big(\sum_{x\in \cS} \sqrt{p(x|s,a)\big(1 - p(x|s,a)\big)}\Big)^2.$$ 

\vspace{-3mm}\noindent
In Lemma \ref{lem:localeffective} below we show that $L_{s,a}$ is always controlled by the number $K_{s,a}$ of successor states of $(s,a)$.\footnote{We recall that for a pair $(s,a)$, we define $\cK_{s,a}:=\supp(p(\cdot|s,a))$, and denote its cardinality by $K_{s,a}$.} The lemma also relates $L_{s,a}$ to the Gini index of the transition distribution of $(s,a)$, defined as $G_{s,a} := \sum_{x\in \cS} p(x|s,a)(1-p(x|s,a))$. 

\begin{lemma}[Local effective support]\label{lem:localeffective}
For any $(s,a)$: 
$$
L_{s,a} \leq K_{s,a}G_{s,a} \leq K_{s,a}-1.
$$
\end{lemma}

\begin{theorem}[Regret of \UCRLnew]\label{thm:regret_UCRLnew}
		With probability higher than $1-4\delta$, uniformly over all $T\ge 3$,
		\als{
			\kR(\mathrm{\UCRLnew}&,T) \leq  c\sqrt{T \log\big(\tfrac{6S^2A\sqrt{T+1}}{\delta}\big)}  \\
			&+ 60DKS^{2/3}A^{2/3}T^{1/3} + \cO\Big(DS^2A\log^2\big(\tfrac{T}{\delta}\big)\Big),
		}	
		with 
		$c=5\sum\nolimits_{s,a} D_s^2L_{s,a} +10\sqrt{SA} + 2D$. Therefore, the regret of \UCRLnew\ asymptotically grows as 
		$$
\cO\bigg(\Big[\sqrt{\sum\nolimits_{s,a} \big(D_s^2L_{s,a}\lor 1\big)} + D\Big]\sqrt{T\log(\sqrt{T}/\delta)}\bigg).
$$		
\end{theorem}
	
We now compare the regret bound of \UCRLnew\ against that of \UCRLB. As shown in Table \ref{table:alg_summary}, the latter algorithm attains a regret bound of $\cO(\sqrt{D\sum_{s,a}K_{s,a} T\log(T)\log(T/\delta)})$. The two regret bounds are not directly comparable: The regret bound of \UCRLB\ depends on $\sqrt{D}$ whereas that of \UCRLnew\ has a term scaling as $D$. However, the regret bound of \UCRLB\ suffers from an additional $\sqrt{\log(T)}$ term. Let us compare the two bounds for MDPs where quantities such as $K_{s,a}$, $L_{s,a}$, and $D_s$ are \emph{local} parameters in the sense that they do not scale with $S$, but where $D$ could grow with $S$ (one example is RiverSwim) --- In other words, $K_{s,a}$, $L_{s,a}$, and $D_s$ scale as $o(S)$. In such a case, comparing the two bounds boils down to comparing $(\sqrt{SA} + D)\sqrt{T\log(T)}$ against $\sqrt{DSAT\log^2(T)}$. When $T\ge \exp\big(\tfrac{(D+\sqrt{SA})^2}{DSA}\big)$ the effect of $\sqrt{\log(T)}$ is not small, and the regret bound of \UCRLnew\ dominates that of \UCRLB. For instance, in 100-state RiverSwim, this happens for all $T\ge 71$. 
It has been left open whether this latter extra factor can be removed.

\section{Numerical Experiments}\label{sec:NumExp}
In this section we provide illustrative numerical experiments that show the benefit of \UCRLnew\ over \UCRL\ and some of its popular variants. Specifically, we compare the empirical performance of \UCRLnew\ against that of state-of-the-art algorithms including  \UCRL, \KLUCRL, and \UCRLB\ --- We also present further results in Appendix E, where we empirically compare \UCRLnew\ against \PSRL. For all algorithms, we set $\delta=0.05$ and use the same tie-breaking rule. 
The full code and implementation details are made available to the community (see Appendix D for details).

\begin{figure}[t]
	\label{fig:river_swim}
	\centering
	\tiny
	\def\svgwidth{1\columnwidth}
	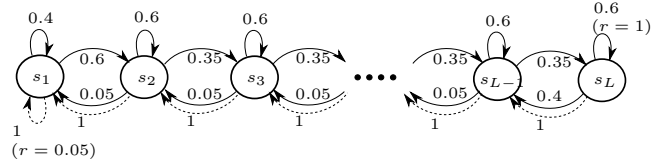
	\vspace{-5mm}
	\caption{The $L$-state \emph{RiverSwim} MDP}
	\vspace{-5mm}
\end{figure}

In the first set of experiments, we consider the $S$-state RiverSwim environment (corresponding to the MDP shown in Figure~\ref{fig:river_swim}). To better understand Theorem~\ref{thm:regret_UCRLnew} in this environment, 
we report in Table~\ref{table:MDP_properties_RiverSwim} a computation of some of the key quantities appearing in the regret bounds, as well as the diameter $D$, for several values of $S$. 
We further provide in Table~\ref{table:Regret_bounds_comparison_RiverSwim} a computation of the leading terms of several regret analyses. More precisely, for a given algorithm $\bA$, we introduce $\overline\kR(\bA)$ to denote the regret bound normalized by $\sqrt{T\log(T/\delta)}$ \emph{ignoring} universal constants. For instance, $\overline\kR(\UCRL) = D\sqrt{SA}$.\footnote{Ignoring universal constants here provides a more fair comparison; for example the final regret bound of \UCRL\ has no second-order term at the expense of a rather large universal constant. Another reason in doing so is that for \UCRLB\ and \SCAL$^+$, universal constants in their corresponding papers are not reported.} In Table \ref{table:Regret_bounds_comparison_RiverSwim}, we compare $\overline \kR$ for various algorithms, for $S$-state RiverSwim for several values of $S$. We stress that $\overline\kR(\UCRLB)$ grows with $T$ unlike $\overline\kR$ for \UCRL, \SCAL$^+$, and \UCRLnew.
Note that even choosing a small value of $T=100$, and ignoring universal constants (which disadvantage \UCRLnew), we get smaller regret bounds with \UCRLnew.

\begin{table}
	\scriptsize
	\centering
	\begin{tabular}[!hbtp]{cccccc}
		\hline
		$S$ & $D$ & $\min_{s}\!D_s$ &  $\max_s\!D_s$ & $\min_{s,a}\!L_{s,a}$ & $\max_{s,a}\!L_{s,a}$\\ \hline
		$6$ & 14.72 & 1.67 & 6.66  & 0  & 1.40 \\
		$12$ & 34.72 & 1.67  & 6.67 & 0 & 1.40\\
		$20$ & 61.39 & 1.67  & 6.67 & 0 & 1.40 \\
		$40$ & 128.06 & 1.67 & 6.67  & 0 & 1.40\\
		$100$ & 328.06 & 1.67 &  6.67 & 0 & 1.40\\ \hline
	\end{tabular}
	\vspace{-2mm}
	\caption{Problem-dependent quantities for $S$-state \emph{RiverSwim}}
	\label{table:MDP_properties_RiverSwim}
	\vspace{-2mm}
	\normalsize
\end{table}

{\color{blue}
	\begin{table}
		\scriptsize
		\centering
		\begin{tabular}[b]{cccccc}
			\hline
			$S$ &  $\overline\kR(\UCRL)$ & $ \overline\kR(\SCAL^+)$ & $\overline\kR(\UCRLB)$ & $\overline\kR(\UCRLnew)$ \\ \hline
			$6$ & $124.9$ & $69.1$ &  $38.6$   & $30.0$ \\
			$12$ & $589.3$ & $235.5$ & $85.8$  & $59.5$ \\
			$20$ & $1736.3$ & $542.2$ & $148.5$  & $94.9$ \\
			$40$ & $7243.9$ & $1609.6$ &  $305.3$ & $176.9$\\
			$70$ & $22576$ & $3802.4$ &   $540.0$ & $293.6$ \\
			$100$ & $46394$ & $6544.7$ &  $775.3$ & $407.6.2$ \\ \hline
		\end{tabular}
		\vspace{-2mm}
		\caption{\small Comparison of the quantity $\overline\kR$ of various algorithms for $S$-state \emph{RiverSwim}: $\overline\kR(\UCRL) = DS\sqrt{A}$, $\overline\kR(\SCAL^+) = D\sqrt{\sum_{s,a} K_{s,a}}$, $\overline\kR(\UCRLB) = \sqrt{D\sum_{s,a}K_{s,a} \log(T)}$ for $T=100$, and $\overline\kR(\UCRLnew) = \sqrt{\sum_{s,a} (D_s^2 L_{s,a}\lor 1)} + D$}
		\vspace{-5mm}
		\label{table:Regret_bounds_comparison_RiverSwim}		
		\normalsize
	\end{table}
}

\begin{figure}[htbp]
	\begin{minipage}[r]{0.5\textwidth}
		\begin{center}
			\includegraphics[width=0.9\linewidth, trim = {0 0 0 10mm}, clip]{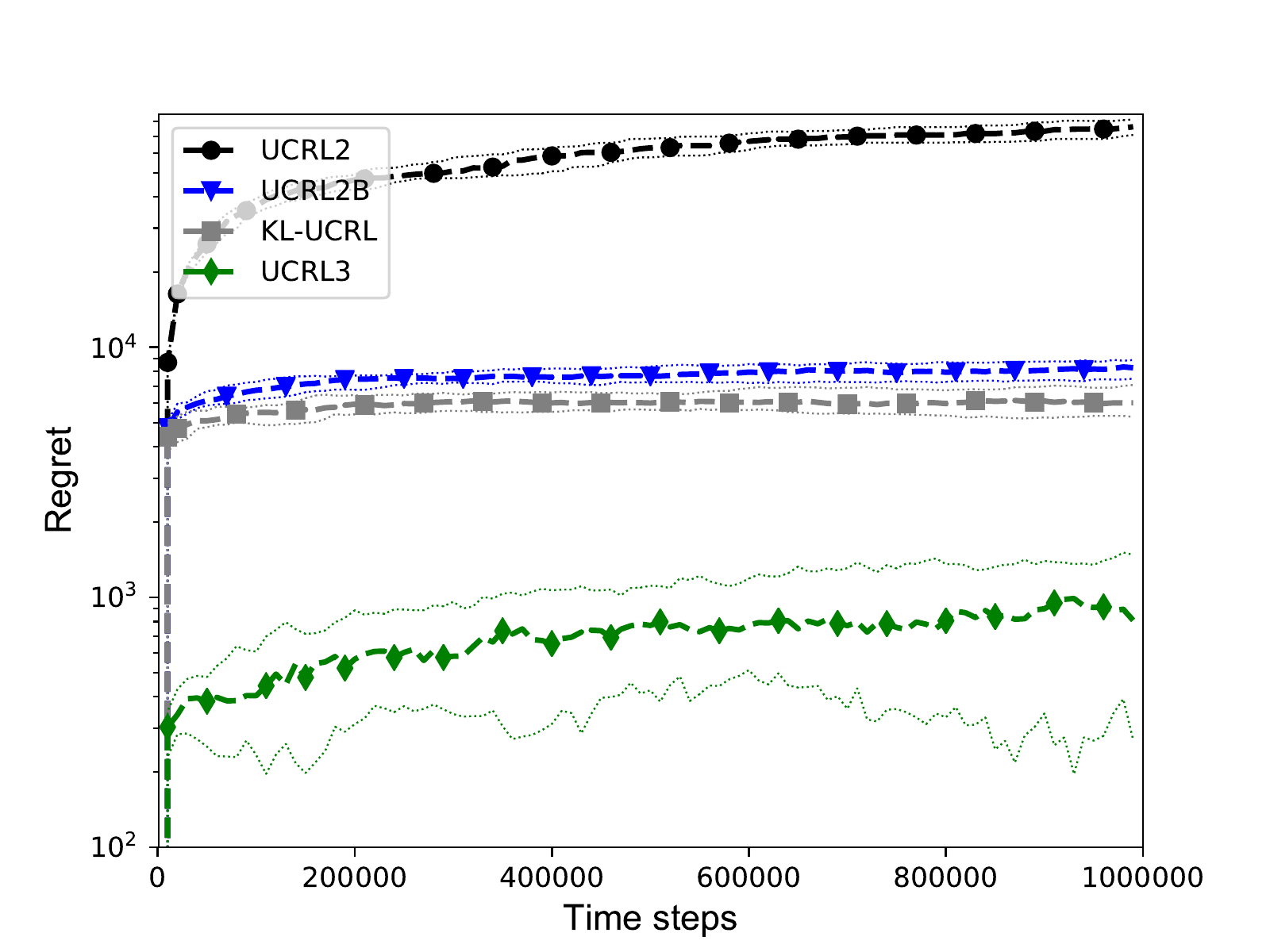}
		\end{center}
	\end{minipage}\hfill
	\begin{minipage}[l]{0.5\textwidth}	
		\begin{center}      	
			\vspace{-3mm}
			\caption{Regret for the 6-state \textit{RiverSwim} environment }   	
			\vspace{-5mm}
			\label{fig:6riverswim}
		\end{center}
	\end{minipage}
\end{figure}

\begin{figure}[htbp]
	\begin{minipage}[r]{0.5\textwidth}
		\begin{center}
			\includegraphics[width=0.9\linewidth, trim = {0 0 0 10mm}, clip]{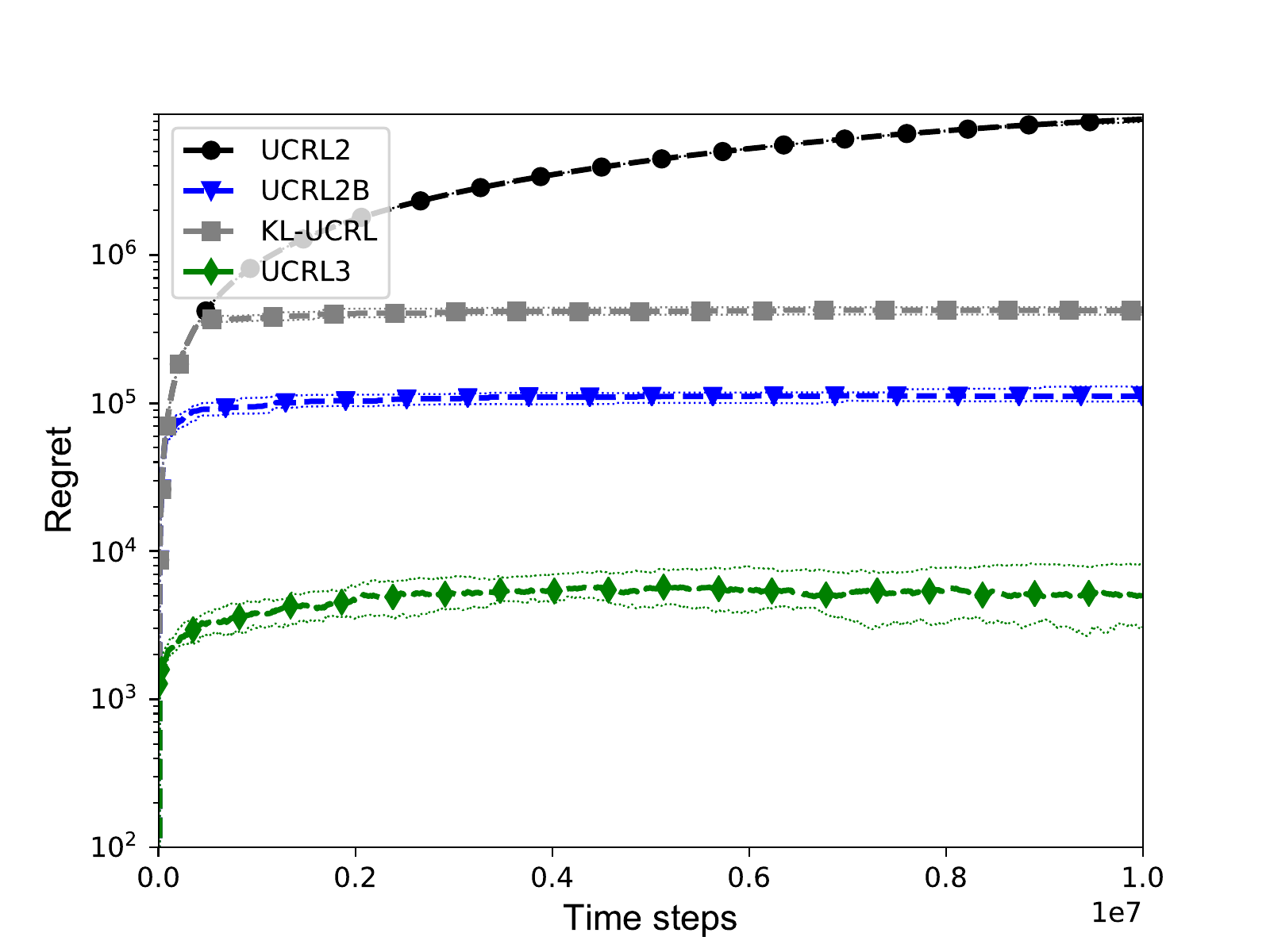}
		\end{center}
	\end{minipage}\hfill
	\begin{minipage}[l]{0.5\textwidth}	
		\begin{center}      	
			\vspace{-3mm}
			\caption{Regret for the 25-state \textit{RiverSwim} environment }   	
			\vspace{-5mm}
			\label{fig:25riverswim}
		\end{center}
	\end{minipage}
\end{figure}

In Figure~\ref{fig:6riverswim}, we plot the regret under \UCRL, \KLUCRL, \UCRLB, and \UCRLnew\ examined in the $6$-state RiverSwim environment.  The curves show the results averaged over $50$ independent runs along with the first and the third quantiles. 
We observe that \UCRLnew\ achieves the smallest regret amongst these algorithms and significantly outperforms \UCRL, \KLUCRL, and \UCRLB\ (note the logarithmic scale).
Figure~\ref{fig:25riverswim} shows similar results on the larger $25$-state RiverSwim environment. 

\vspace{-3mm}

\begin{figure}[h]
	\begin{minipage}[r]{0.5\textwidth}
		\begin{center}
			\includegraphics[height=0.5\linewidth,trim = {32mm 0 32mm 10mm}, clip]{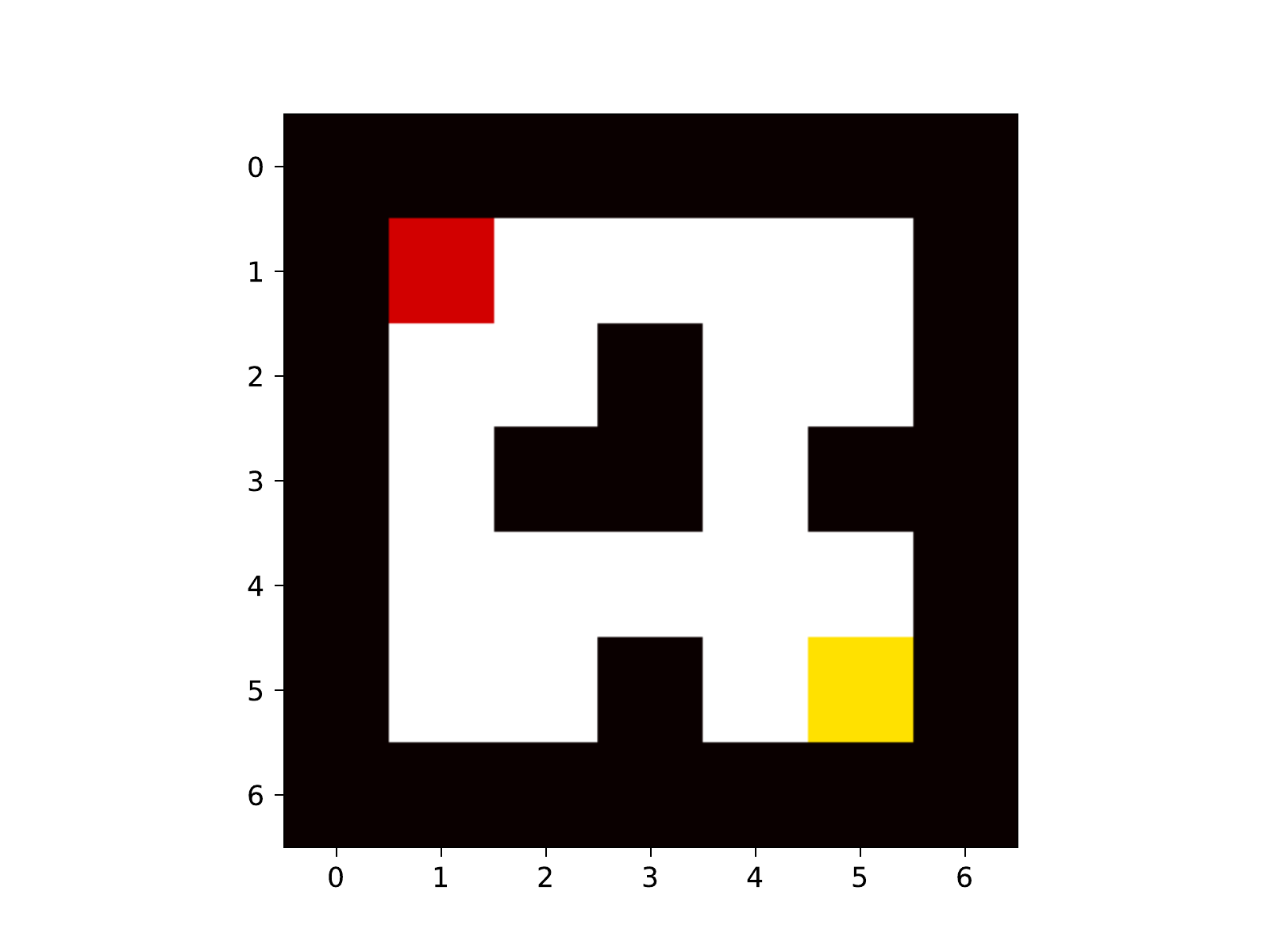}\hfill		
			\includegraphics[height=0.5\linewidth,trim = {20mm 0 22mm 10mm}, clip]{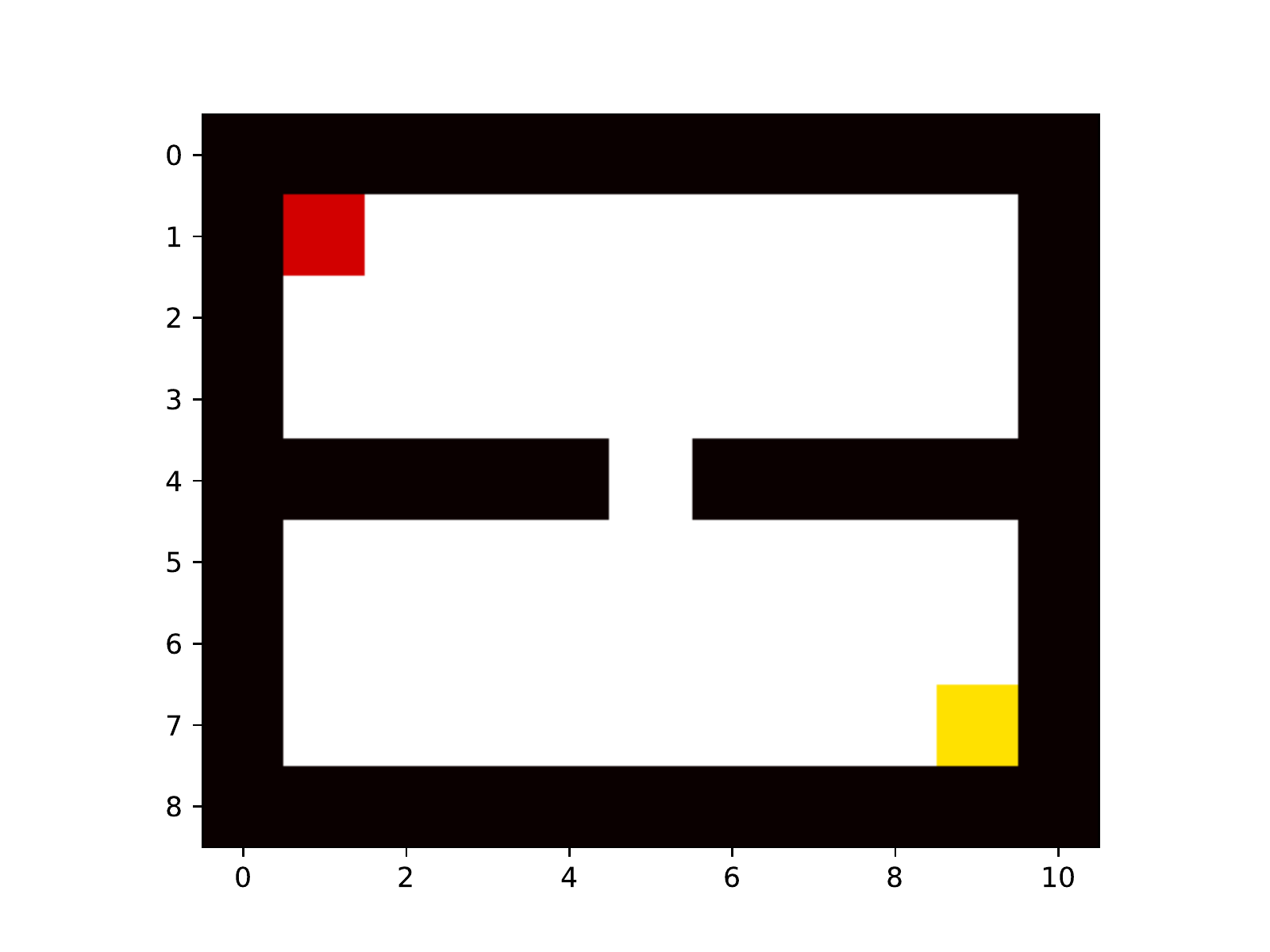}
		\end{center}
	\end{minipage}\hfill
	\begin{minipage}[l]{0.5\textwidth}	
		\begin{center}      	
			\vspace{-2mm}
			\caption{A 4-room (left) and a 2-room (right) grid-world environment, with $20$ and $55$ states: the starting state is shown in red, and the rewarding state is shown in yellow. From the yellow state, all actions bring the learner to the red state. Other transitions are noisy as in a \textit{frozen-lake} environment.}
			\vspace{-3mm}
			\label{fig:4R_2R_environments}
		\end{center}
	\end{minipage}
\end{figure}

We further provide results in larger MDPs. We consider two frozen lake environments of respective sizes  of $7 \times 7$  and $9\times 11$ as shown in Figure~\ref{fig:4R_2R_environments}, thus yielding MDPs with, respectively, $S=20$ and $S=55$ states (after removing walls). 
In such grid-worlds, the learner starts in the upper-left corner. A reward of $1$ is placed in the lower-right corner, and the rest of states give no reward. Upon reaching the rewarding state, the learner is sent back to the initial state. 
The learner can perform 4 actions (when away from walls): Going up, left, down, or right. Under each, the learner moves in the chosen direction (with probability $0.7$), stays in the same state (with probability $0.1$), or goes in each of the two perpendicular directions (each with probability $0.1$) -- Walls act as reflectors moving back the leaner to the current state. 

\begin{remark}
	Importantly, \UCRL\ and its variants are generic purpose algorithms, and as such, are not aware of the specific structure of the MDP, such as being a grid-world. In particular, no prior knowledge is assumed on the support of the transition distributions by any of the algorithms, which makes it a highly non-trivial learning task, since the number of unknowns (i.e., problem dimension) is then $S^2A$ ($SA(S-1)$ for the transition function, and $SA$ for the rewards). For instance, a 4-room MDP is really seen as a problem of dimension $1600$ by these algorithms, and a 2-room MDP as a problem of dimension $12100$.
\end{remark}

\vspace{-3mm}

\begin{figure}[htbp]   	
	\vspace{-3mm}
	\begin{minipage}[r]{0.5\textwidth}
		\begin{center}
			\includegraphics[width=0.9\linewidth, trim = {0 0 0 10mm}, clip]{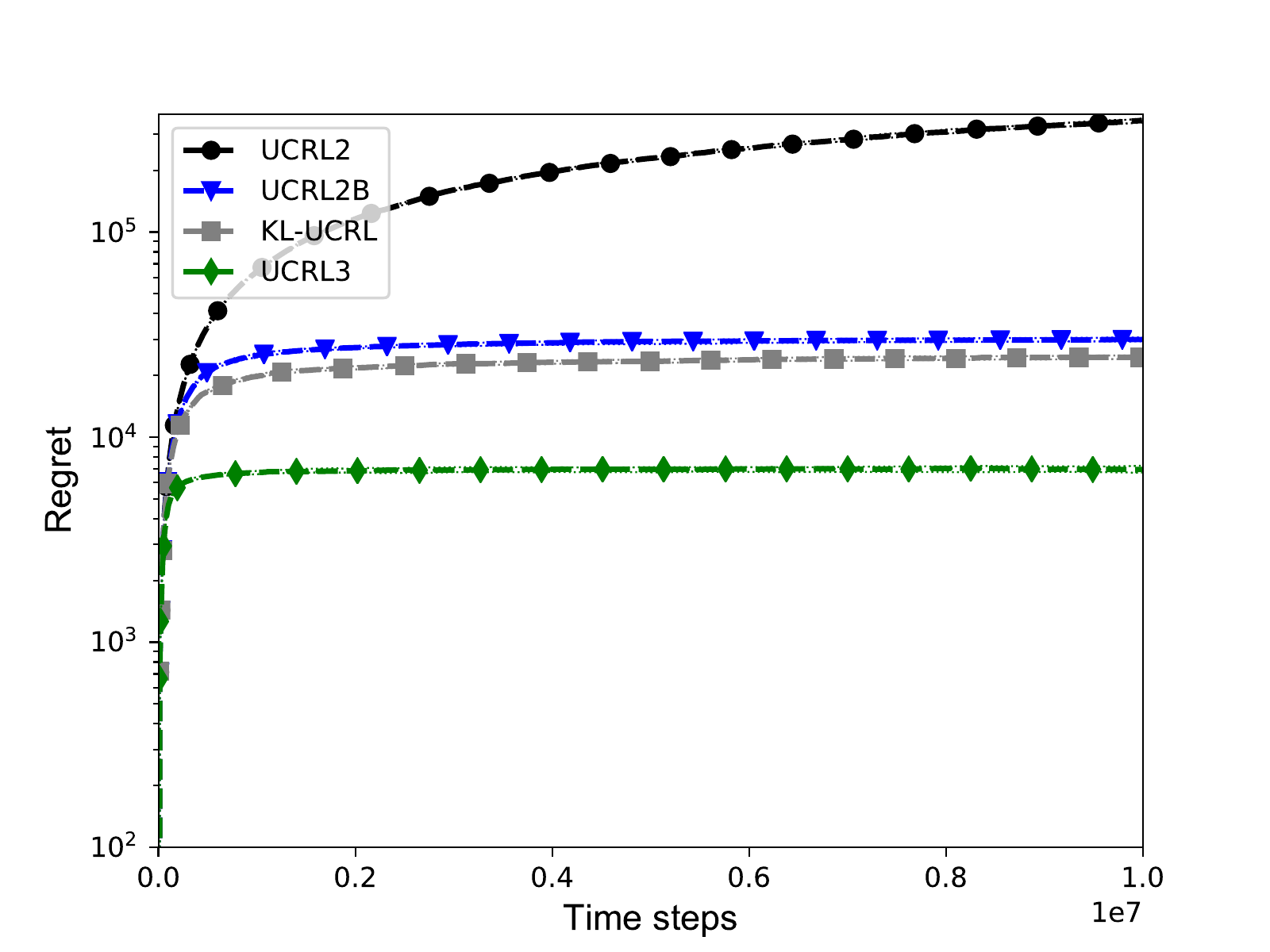}\\
		\end{center}
	\end{minipage}\hfill
	\begin{minipage}[l]{0.5\textwidth}	
		\begin{center}      	
			\vspace{-3mm}
			\caption{Regret for the 4-room environment }
			\vspace{-8mm}
			\label{fig:4room}
		\end{center}
	\end{minipage}
\end{figure}

\begin{figure}[htbp]   	
	\vspace{-2mm}
	\begin{minipage}[r]{0.5\textwidth}
		\begin{center}
			\includegraphics[width=0.9\linewidth, trim = {0 0 0 10mm}, clip]{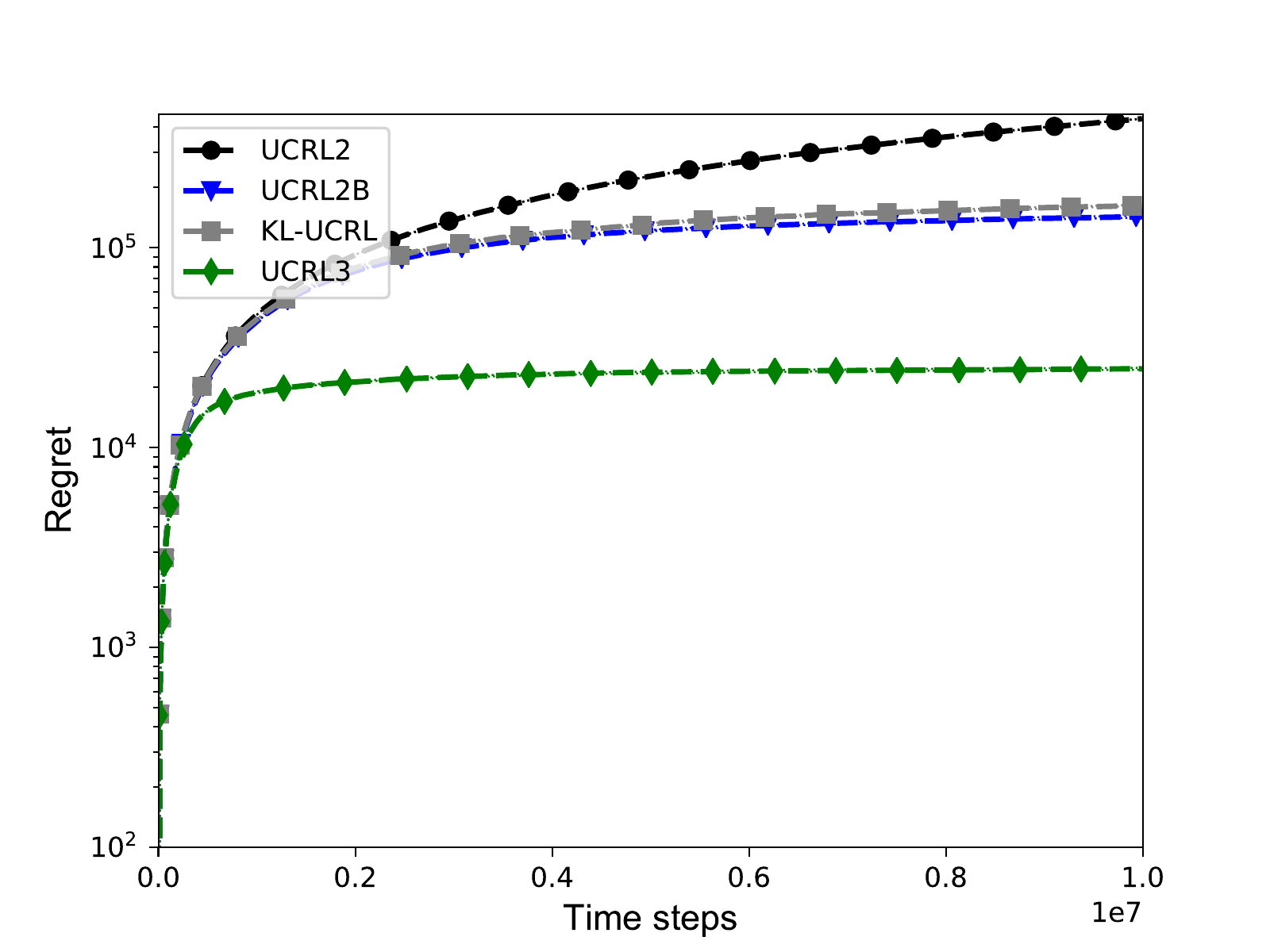}\\
		\end{center}
	\end{minipage}\hfill
	\begin{minipage}[l]{0.5\textwidth}	
		\begin{center}      	
			\vspace{-3mm}
			\caption{Regret for the 2-room environment }
			\vspace{-5mm}
			\label{fig:2room}
		\end{center}
	\end{minipage}
\end{figure}

Figures~\ref{fig:4room} (respectively, Figure \ref{fig:2room}) shows the regret performance of  \UCRL, \KLUCRL, \UCRLB, and \UCRLnew\ in the 
2-room (respectively, 4-room) grid-world MDP. Finally, since all these algorithms are generic-purpose MDP learners, we provide numerical experiments in a large randomly-generated MDP
consisting of $100$ states and $3$ actions, hence seen as being of dimension $3\times 10^4$.  \UCRLnew\ still outperforms other state-of-the-art algorithms by a large margin consistently in all these environments. We provide below, an illustration of a randomly-generated MDP, with $15$ states and $3$ actions (blue, red, green).
Such an MDP is a type of Garnet (Generalized Average Reward Non-stationary Environment Test-bench) introduced in \cite{bhatnagar2009natural}, in which we can specify the numbers of states and actions, the average size of the support of transition distributions, the sparsity of the reward function,
as well as the minimal non-zero probability mass and minimal non-zero mean-reward.

Comparing \UCRLnew\ against \UCRLB\ in experiments reveals that the gain achieved here is not only due to Bernstein's confidence intervals. 
Let us recall that on top of using  Berstein's confidence intervals, \UCRLnew\ 
also uses a refinement using sub-Gaussianity of Bernoulli distributions as well as the \texttt{EVI-NOSS} instead of \EVI\ for planning. Experimental results verify that both tight confidence sets (see also 
Figure 11 
in the appendix) and \texttt{EVI-NOSS} play an essential role in achieving small empirical regret.

\begin{figure}[hbtp]   	
	\vspace{-1mm}
	\begin{center}
		\includegraphics[width=\linewidth, trim = {0cm 1.6cm 0 2.cm},clip]{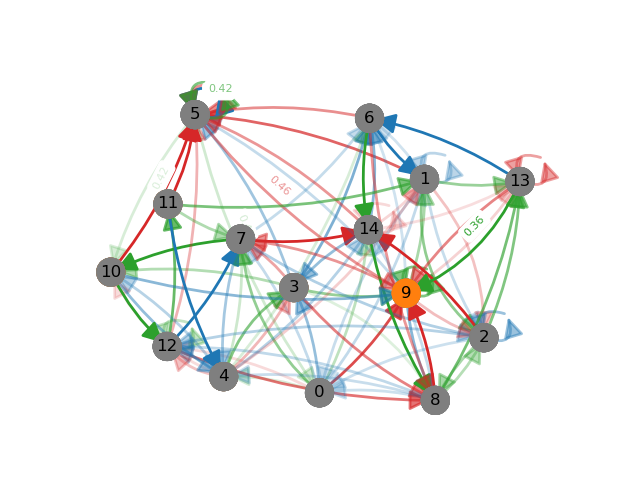}
	\end{center}
	\vspace{-5mm}
	\caption{A randomly-generated MDP with $15$ states: One color per action, shaded according to the corresponding probability mass, labels indicate mean reward, and the current state is highlighted in orange.}\label{fig:15randomMDP}
\end{figure}

\begin{figure}[htbp]   	
	\vspace{-3mm}
	\begin{minipage}[r]{0.5\textwidth}
		\begin{center}
			\includegraphics[width=0.9\linewidth, trim = {0 0 0 10mm}, clip]{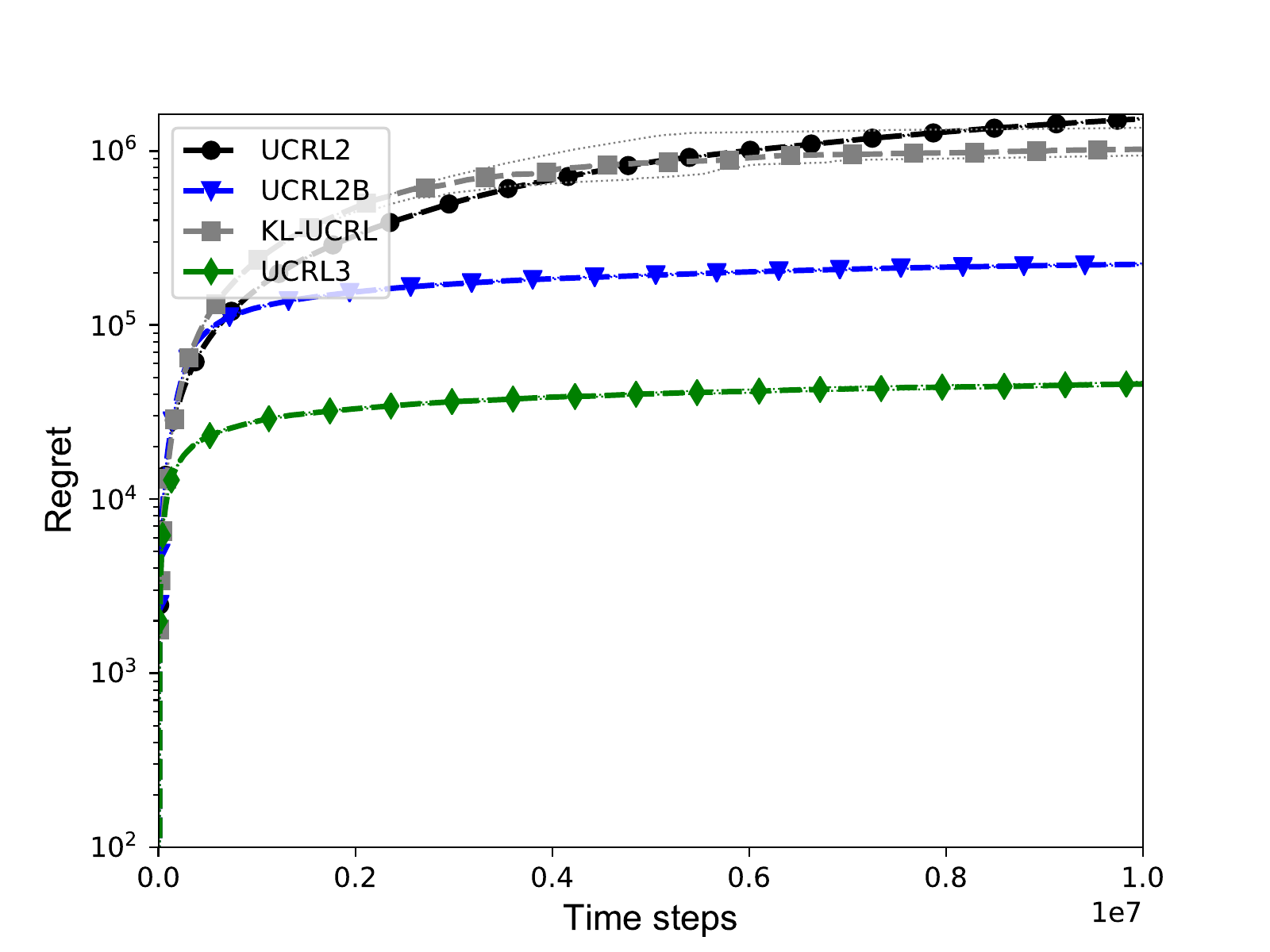}\\
		\end{center}
	\end{minipage}\hfill
	\begin{minipage}[l]{0.5\textwidth}	
		\begin{center}      	
			\vspace{-3mm}
			\caption{Regret in one $100$-state randomly generated MDP}
			\vspace{-5mm}
			\label{fig:randomMDP}
		\end{center}
	\end{minipage}
\end{figure}

\section{Conclusion}
We studied reinforcement learning in finite Markov decision processes (MDPs) under the average-reward criterion, and introduced  \UCRLnew, a refined variant of \UCRL\ \citep{jaksch2010near}, that efficiently balances exploration and exploitation in communicating MDPs. The design of \UCRLnew\ combines two main ingredients: (i) Tight time-uniform confidence bounds on individual elements of transition and reward functions, and (ii) a refined Extended Value Iteration procedure being adaptive to the support of transition function. 
We provided a non-asymptotic and high-probability regret bound for \UCRLnew\ scaling as $\widetilde \cO\big(\big(D+\sqrt{\sum_{s,a}(D_s^2L_{s,a}\lor 1)}\big)\sqrt{T}\big)$, where $D$ denotes the (global) diameter of the MDP, $D_s$ denotes the \emph{local} diameter of state $s$, and $L_{s,a}$ represents the local effective support of transition distribution for state-action pair $(s,a)$. We further showed that $D_s\le D$ and that $L_{s,a}$ is upper bounded by the number of successor states of $(s,a)$, and therefore, the above regret bound improves on that of \UCRL. Through numerical experiments we showed that \UCRLnew\ significantly outperforms existing variants of \UCRL\ in standard environments. An interesting yet challenging research direction is to derive problem-dependent logarithmic regret bounds for \UCRLnew. 

\section*{Acknowledgement}
This work has been supported by CPER Nord-Pas-de-Calais/FEDER DATA Advanced data science
and technologies 2015-2020, the French Ministry of Higher Education and Research, Inria, and
the French Agence Nationale de la Recherche (ANR), under grant ANR-16-CE40-0002 (the BADASS project). Part of this work was done while M.~S.~Talebi was a postdoctoral researcher in Inria Lille -- Nord Europe.

\bibliography{Bandit_RL_bib,2018library}
\bibliographystyle{icml2020}


\appendix
\onecolumn
\section{Concentration Inequalities}
\label{app:concentration}

\subsection{Time-Uniform Laplace Concentration for Sub-Gaussian Distributions}
\begin{definition}[Sub-Gaussian observation noise]\label{def:subGaussian}
    A sequence $(Y_t)_{t}$ has conditionally $\sigma$-sub-Gaussian noise if	
	\beqan
	\forall t, \forall \lambda\in\Real,\quad  \log \Esp[ \exp\big(\lambda (Y_t- \Esp[Y_t|\cF_{t-1}])\big) \big| \cF_{t-1}] \leq \frac{\lambda^2 \sigma^2}{2}\,,
	\eeqan
    where $\cF_{t-1}$ denotes the $\sigma$-algebra generated by $Y_1,\ldots,Y_{t-1}$.
\end{definition}

\begin{lemma}[Uniform confidence intervals]\label{lem:confintervals}
	Let $Y_1,\ldots, Y_t$ be a sequence of $t$ i.i.d.~real-valued random variables with mean $\mu$, such that $Y_t-\mu$ is  $\sigma$-sub-Gaussian. Let
	$\mu_t= \frac{1}{t}\sum_{s=1}^t Y_s$ be the empirical mean estimate. Then, for all $\delta\in(0,1)$, it holds
	\beqan
	\Pr\bigg(\exists t\in\Nat,\quad |\mu_t -\mu| \geq \sigma\sqrt{\Big(1+\frac{1}{t}\Big)\frac{2\ln\big(\sqrt{t+1}/\delta\big)}{t}}\bigg)& \leq& \delta\,.
	\eeqan		
\end{lemma}

The ``Laplace" method refers to using the Laplace method of integration for optimization.

\begin{myproof}{of Lemma~\ref{lem:confintervals}}
	We introduce for a fixed $\delta\in(0,1)$ the random variable
	\beqan
	\tau = \min\bigg\{  t\in\Nat :\mu_t -\mu  \geq   \sigma\sqrt{\Big(1+\frac{1}{t}\Big)\frac{2\ln\big(\sqrt{1+t}/\delta\big)}{t}} \bigg\}\,.
	\eeqan
	This quantity is a random stopping time for the filtration $\cF = (\cF_{t})_{t}$, where $\cF_t = \sigma(Y_1,\dots,Y_t)$, since  $\{ \tau \leq m\}$ is $\cF_m$-measurable for all $m$. We want to show that
	$\Pr(\tau<\infty) \leq \delta$. To this end, for any $\lambda$ and $t$, we introduce the following quantity:
	\beqan
	M^\lambda_t = \exp\bigg(\sum_{s=1}^t \Big(\lambda (Y_s-\mu) - \frac{\lambda^2\sigma^2}{2}\Big)\bigg).
	\eeqan
	By assumption, the centered random variables are $\sigma$-sub-Gaussian and it is immediate to show that $(M^\lambda_t)_{t\in\Nat}$ is a non-negative super-martingale that satisfies $\ln \Esp[M^\lambda_t] \leq 0$ for all $t$. 	It then follows that  $M^\lambda_\infty = \lim_{t\to\infty} M_{t}^\lambda$ is almost surely well-defined and  so is $M_{\tau}^\lambda$.
	Furthermore, using the fact that  $M_t^\lambda$ and $\{\tau>t\}$ are $\cF_t$-measurable, it comes
	\beqan
	\Esp[M_\tau^\lambda] &=& \Esp[M_1^\lambda] +  \Esp[\sum_{t=1}^{\tau-1} M^\lambda_{t+1}-M_t^\lambda]\\
	&=& 1 + \sum_{t=1}^\infty \Esp[(M_{t+1}^\lambda-M_t^\lambda)\indic{\tau>t}]\\
	&=& 1 + \sum_{t=1}^\infty \Esp[(\Esp[M_{t+1}^\lambda|\cF_t]-M_t^\lambda)\indic{\tau>t}]\\
	&\leq& 1\,.
	\eeqan

	The next step is to introduce the auxiliary variable $\Lambda\sim \cN(0,\sigma^{-2})$, independent of all other variables, and study the quantity $M_t = \Esp[M_t^\Lambda|\cF_\infty]$.  Note that the standard deviation of $\Lambda$ is $\sigma^{-1}$ due to the fact we consider $\sigma$-sub-Gaussian random variables.
	We immediately get $\Esp[M_\tau]= \Esp[\Esp[M_\tau^\Lambda |\Lambda]]\leq 1$.
	For convenience, let $S_t=t(\mu_t-\mu)$. By construction of $M_t$, we have
	\beqan
	M_t&=& \frac{1}{\sqrt{2\pi\sigma^{-2}}}\int_\Real \exp\bigg(  \lambda S_t  -  \frac{\lambda^2\sigma^2 t}{2} - \frac{\lambda^2\sigma^2}{2}\bigg)\mathrm{d}\lambda\\
	&=& \frac{1}{\sqrt{2\pi\sigma^{-2}}}\int_\Real \exp\bigg( - \bigg[\lambda \sigma \sqrt{\frac{t+1}{2}} - \frac{S_t}{\sigma\sqrt{2(t+1)}} \bigg]^2
	+ \frac{S_t^2}{2\sigma^2(t+1)}\bigg)\mathrm{d}\lambda\\
	&=&\exp\bigg(\frac{S_t^2}{2\sigma^2(t+1)}\bigg)\frac{1}{\sqrt{2\pi\sigma^{-2}}}\int_\Real \exp\Big(-\lambda^2 \sigma^2\frac{t+1}{2}\Big)\mathrm{d}\lambda\\
	&=&\exp\bigg(\frac{S_t^2}{2\sigma^2(t+1)}\bigg)\frac{\sqrt{2\pi\sigma^{-2}/ (t+1) }}{\sqrt{2\pi\sigma^{-2}}}\,.
	\eeqan
	Thus, we deduce that	
	\beqan
	|S_t|=  \sigma\sqrt{2(t+1)\ln\big(\sqrt{t+1}M_t\big)}\,.
	\eeqan
	We conclude by applying a simple Markov's inequality:
	\beqan
	\Pr\bigg( \tau|\mu_\tau-\mu| \geq  \sigma\sqrt{2(\tau+1)\ln\big(\sqrt{\tau+1}/\delta\big)} \bigg) =
	\Pr( M_\tau \geq 1/\delta ) \leq \Esp[M_\tau]\delta\,.
	\eeqan
\end{myproof}

\subsection{Time-Uniform Laplace Concentration for Bernoulli Distributions}\label{app:Bernoulliconcentration}
We now want to make use of  the special structure of Bernoulli variables to derive refined time-uniform concentration inequalities. Let us first recall that if $(X_i)_{i\leq n}$ are i.i.d.~according to a Bernoulli distribution $\cB(p)$ with parameter $p\in[0,1]$, then it holds by the Chernoff-method that for all $\epsilon\geq 0$,
\als{
	\Pr\bigg(\frac{1}{n}\sum_{i=1}^n(X_i - p) \geq \epsilon\bigg) \leq \exp\bigg(-n\kl(p+\epsilon,p)\bigg)\,,
}
where $\kl(p,q)=p\log(p/q)+(1-p)\log((1-p)/(1-q))$ denotes the Bernoulli Kullback-Leibler divergence. The reverse map of the Cram\'er transform $\epsilon\mapsto \kl(p+\epsilon,p)$ is unfortunately not explicit, and one may consider Taylor's approximation of it to derive approximate but explicit high-probability confidence bounds.
More precisely, the following has been shown (see \citep{kearns1998large,weissman2003inequalities,berend2013concentration,raginsky2013concentration}):
\begin{lemma}[Sub-Gaussianity of Bernoulli random variables]
	For all  $p\in[0,1]$, the left and right tails of the Bernoulli distribution are controlled in the following way
	\beqan
	\forall \lambda \in\Real, &&
	\log \Esp_{X\sim\cB(p)}\big[\exp(\lambda (X-p))\big] \leq \frac{\lambda^2}{2} g(p) \,,
	\eeqan
	where $g(p) = \frac{1/2-p}{\log(1/p-1)}$. The control of the right-tail can be further refined for $p\in[\frac{1}{2},1]$ as follows:
	\beqan
	\forall \lambda \in\Real^+, &&
	\log \Esp_{X\sim\cB(p)}\big[ \exp(\lambda (X-p))\big] \leq \frac{\lambda^2}{2} p(1-p)\,.
	\eeqan
\end{lemma}
We note that the left and right tails are not controlled in a symmetric way. This yields, introducing the function $\underline{g}(p) =\begin{cases} g(p)&\text{if }p<1/2\\ p(1-p) &\text{otherwise}\end{cases}$, the following asymmetrical  confidence set
\begin{corollary}[Time-uniform Bernoulli concentration]\label{cor:timeuniformBernoulliconcentration}
	Let $\!(X_i)_{i\leq n}\!\!\stackrel{\text{i.i.d.}}{\sim}\!\!\cB(p)$. Then, for all $\delta\!\in\!(0,1)$,	
	\beqan
	\!\!	\Pr\bigg(\!\forall n\!\in\!\Nat,\,
	-\!\sqrt{g(p)}\beta_n(\delta) \leq \!\frac{1}{n}\! \sum_{i=1}^n X_i \!-\! p \leq\! \sqrt{\underline{g}(p)}\beta_n(\delta)
	\bigg)\!\geq\! 1\!-\!2\delta\,,
	\eeqan
	where $\beta_n(\delta):= \sqrt{\frac{2}{n} \big(1+\frac{1}{n}\big)\log(\sqrt{n+1}/\delta)}$.
\end{corollary}

\begin{myproof}{of Corollary~\ref{cor:timeuniformBernoulliconcentration}}
	Let us introduce the following quantities
	\beqan
	\forall \lambda \in\Real^+,\quad M^\lambda_t &=& \exp\bigg(\sum_{s=1}^t \Big(\lambda (X_s-p) - \frac{\lambda^2\underline{g}(p)}{2}\Big)\bigg)\,,\\
	\forall \lambda \in\Real, \quad M'^\lambda_t &=& \exp\bigg(\sum_{s=1}^t \Big(\lambda (X_s-p) - \frac{\lambda^2g(p)}{2}\Big)\bigg)\,.
	\eeqan
	Note that $M^\lambda_t$ is a non-negative super-martingale for all $\lambda\in\Real^+$,
	and $M'^\lambda_t$ is a  non-negative super-martingale for all $\lambda\in\Real$.
	Furthermore, $\Esp[M_t^\lambda]\leq1$ and $\Esp[{M'}_t^\lambda]\leq 1$.

	Let $\Lambda$ be a random variable with density
	\beqan
	f_p(\lambda)= \begin{cases}
		\frac{\exp(-\lambda^2\underline{g}(p)/2)}{\int_{\Real^+} \exp(-z^2\underline{g}(p)/2)\mathrm{d}z} = \sqrt{\frac{2\underline{g}(p)}{\pi}}\exp(-\lambda^2\underline{g}(p)/2)& \text{ if }\lambda\in\Real^+,\\
		0& \text{ else}.
	\end{cases}
	\eeqan
	
	Let $M_t = \Esp[M_t^\Lambda|\cF_t]$ and note that
	\beqan
	M_t&=&  \sqrt{\frac{2\underline{g}(p)}{\pi}}\int_{\Real^+} \exp\bigg(  \lambda S_t  -  \frac{\lambda^2\underline{g}(p) t}{2} - \frac{\lambda^2\underline{g}(p)}{2}\bigg)\mathrm{d}\lambda\\
	&=& \sqrt{\frac{2\underline{g}(p)}{\pi}}\int_{\Real^+} \exp\bigg( - \Big[\lambda  \sqrt{\frac{\underline{g}(p)(t+1)}{2}} - \frac{S_t}{\sqrt{2\underline{g}(p)(t+1)}} \Big]^2
	+ \frac{S_t^2}{2\underline{g}(p)(t+1)}\bigg)\mathrm{d}\lambda\\
	&=&\exp\bigg(\frac{S_t^2}{2\underline{g}(p)(t+1)}\bigg)\sqrt{\frac{2\underline{g}(p)}{\pi}}\int_{\Real^+} \exp\Big(-\Big(\lambda - \frac{S_t}{\underline{g}(p)(t+1)}\Big)^2 \underline{g}(p)\frac{t+1}{2}\Big)\mathrm{d}\lambda\\
	&=&\exp\bigg(\frac{S_t^2}{2\underline{g}(p)(t+1)}\bigg)\sqrt{\frac{2\underline{g}(p)}{\pi}}\int_{c_t} \exp\Big(-\lambda^2 \underline{g}(p)\frac{t+1}{2}\Big)\mathrm{d}\lambda \qquad \text{ where } c_t = - \frac{S_t}{\underline{g}(p)(t+1)}\\
	&\geq&\exp\bigg(\frac{S_t^2}{2\underline{g}(p)(t+1)}\bigg)\sqrt{\frac{2\underline{g}(p)}{\pi}}\sqrt{\frac{\pi}{2(t+1)\underline{g}(p)}} \qquad \text{ if }S_t\geq 0\\
	&=&\exp\bigg(\frac{S_t^2}{2\underline{g}(p)(t+1)}\bigg)\frac{1}{\sqrt{t+1}}\, .
	\eeqan
	Note also that $M_t$ is still a non-negative super-martingale satisfying  $\Esp[M_t] \leq 1$ for all $t$.
	Likewise, considering $\Lambda'$ to be a random variable with density
	\beqan
	f'_p(\lambda)= \begin{cases}
		\frac{\exp(-\lambda^2g(p)/2)}{\int_{\Real^-} \exp(-z^2g(p)/2) \mathrm{d}z} = \sqrt{\frac{2g(p)}{\pi}}\exp(-\lambda^2g(p)/2)& \text{ if }\lambda\in\Real^-,\\
		0& \text{ else}.
	\end{cases}
	\eeqan
	Introducing $M'_t = \Esp[{M'}_t^{\Lambda'}|\cF_t]$, it comes
	\beqan
	M'_t&\geq&\exp\bigg(\frac{S_t^2}{2g(p)(t+1)}\bigg)\frac{1}{\sqrt{t+1}} \quad \text{if } S_t\leq 0.
	\eeqan
	$M'_t$ is a non-negative super-martingale satisfying  $\Esp[M_t] \leq 1$ for all $t$.
	Thus, we deduce that	
	\beqan
	\frac{|S_t|}{t}\leq \begin{cases}
		\sqrt{2\underline{g}(p)\frac{(1+1/t)}{t}\ln\big(M_t\sqrt{1+ t} \big)}& \text{ if }  S_t\geq 0\\
		\sqrt{2g(p)\frac{(1+1/t)}{t}\ln\big(M'_t\sqrt{1+ t} \big)} & \text{ if } S_t\leq 0\,,
	\end{cases}
	\eeqan
	which implies
	\beqan
	- \sqrt{2g(p)\frac{(1+1/t)}{t}\ln\big(M'_t\sqrt{1+ t} \big)} \leq \frac{S_t}{t} \leq \sqrt{2\underline{g}(p)\frac{(1+1/t)}{t}\ln\big(M_t\sqrt{1+ t} \big)}\, .
	\eeqan
	
	Combining the previous steps, we thus obtain for each $\delta \in(0,1)$,
	\beqan
	\lefteqn{\hspace{-3cm}
		\Pr\bigg(\exists t,\frac{S_t}{t} \geq \sqrt{2\underline{g}(p)\frac{(1+1/t)}{t}\ln\big(\sqrt{1+ t}/\delta \big)}
		\text{ or }\frac{S_t}{t} \leq - \sqrt{2g(p)\frac{(1+1/t)}{t}\ln\big(\sqrt{1+ t}/\delta \big)}
		\bigg)}\\
	&\leq&
	\Pr\bigg(\exists t, M_t \geq 1/\delta \text{ or } M'_t \geq 1/\delta \bigg)\\
	&\leq& \Pr\big(\exists t, M_t \geq 1/\delta\big) + \Pr\big(\exists t, M'_t \geq 1/\delta\big)\\
	&\leq& \delta(\Esp[\max _t M_t] + \Esp[\max_t M'_t])\\
	&\leq& 2\delta\,.
	\eeqan
	The last inequality holds by an application of  Doob's property for non-negative super-martingales, and using that $\Esp[M_1]=\Esp[M'_1]=1$.
\end{myproof}

\subsection{Comparison of Time-Uniform Concentration Bounds}\label{app:compareconcentration}
In this section, we give additional details about the concentration inequalities used to derive the confidence bounds in \UCRLnew. We first present the following result from \citep{maillard2019mathematics}, which makes use of a generic peeling approach:

\begin{lemma}[{\citep[Lemma~2.4]{maillard2019mathematics}}]
\label{lem:time_peeling}
Let $Z = (Z_t)_{t\in \NN}$ be a sequence of random variables generated by a predictable process, and $\cF=(\cF_t)_{t}$ be its natural filtration. Let $\phi:\RR\to \RR_+$ be a convex upper-envelope of the cumulant generating function of the conditional distributions with $\phi(0)=0$, and let $\phi_\star$ denote its Legendre-Fenchel transform, that is:
\als{
\forall \lambda\in \cD, \forall t, \qquad &\log\EE\left[\exp\big(\lambda Z_t\big)|\cF_{t-1}\right] \le \phi(\lambda)\, , \\
\forall x\in \RR, \qquad &\phi_\star(x) = \sup_{\lambda\in \RR} (\lambda x - \phi(\lambda))\, ,
}
where $\cD = \{\lambda\in \RR: \forall t, \log\EE\left[\exp(\lambda Z_t)|\cF_{t-1}\right] \le \phi(\lambda) <\infty\}$. Assume that $\cD$ contains an open neighborhood of $0$.
Let  $\phi_{\star,+}^{-1}:\Real\to\Real_+$ (resp.~$\phi_{\star,-}^{-1}$) be its reverse map on $\Real_+$ (resp.~$\Real_-$), that is
\als{
\phi^{-1}_{\star,-}(z):=\sup\{x\le 0: \phi_\star(x)>z\} \quad \hbox{and} \quad \phi^{-1}_{\star,+}(z):=\inf\{x\ge 0: \phi_\star(x)>z\}\, .
}
Let $N_n$ be a stopping time that is $\cF$-measurable and almost surely bounded by $n$. Then, for all $\eta\in (1,n]$ and $\delta\in (0,1)$,
\als{
\PP\bigg[\frac{1}{N_n}\sum_{t=1}^{N_n}Z_t \geq  \phi_{\star,+}^{-1}\left(\frac{\eta}{N_n}\log\Big(\left\lceil\frac{\log(n)}{\log(\eta)}\right\rceil\frac{1}{\delta}\Big)\right) \bigg] &\leq \delta\, ,\\
\PP\bigg[\frac{1}{N_n}\sum_{t=1}^{N_n}Z_t \leq  \phi_{\star,-}^{-1}\left(\frac{\eta}{N_n}\log\Big(\left\lceil\frac{\log(n)}{\log(\eta)}\right\rceil\frac{1}{\delta}\Big)\right) \bigg] &\leq \delta\,  .
}
Moreover, if $N$ is a possibly unbounded stopping time that is $\cF$-measurable, then for all $\eta>1$ and $\delta\in (0,1)$,
\als{
\PP\bigg[\frac{1}{N}\sum_{t=1}^{N}Z_t \geq  \phi_{\star,+}^{-1}\left(\frac{\eta}{N}\log\bigg[\frac{\log(N)\log(\eta N)}{\delta\log^2(\eta)}\bigg]\right) \bigg] &\leq \delta\, ,\\
\PP\bigg[\frac{1}{N}\sum_{t=1}^{N}Z_t \leq  \phi_{\star,-}^{-1}\left(\frac{\eta}{N}\log\bigg[\frac{\log(N)\log(\eta N)}{\delta\log^2(\eta)}\bigg]\right) \bigg] &\leq \delta\,  .
}
\end{lemma}


In order to derive the confidence intervals for individual elements $p(s'|s,a), (s,a,s')\in \cS\times \cA\times \cS$ of transition function, we directly apply the above lemma to sub-Gamma random variables. Let us first recall that sub-Gamma random variables satisfy
$
\phi(\lambda) \leq \frac{\lambda^2 v}{2(1-b\lambda)}$, for all $\lambda\in(0,1/b)$; see, e.g., \citep[Chapter~2.4]{boucheron2013concentration}. Therefore,
\als{
    \phi_{\star,+}^{-1}(z) = \sqrt{2v z} + bz   \quad \hbox{and} \quad
    \phi_{\star,-}^{-1}(z) = -\sqrt{2v z} - bz    \, .
}
We finally note that for a Bernoulli distributed random variable with parameter $q$, we have $v=q(1-q)$ and $b=1$.

\citet{dann2017unifying} introduce an alternative time-uniform Bernstein bound.
In order to compare the methods, we introduce the following two functions
\beqa
\label{eq:CB_bern_peeling}
C^{\text{Bernstein-D}}(p,n,\delta) &=&p+ \sqrt{\frac{2p}{n} \ell\!\ell_n(\delta)} + \frac{\ell\!\ell_n(\delta)}{n}\\
&\text{where}& \ell\!\ell_n(\delta) = 2\log\log(\max(e,n)) + \log(3/\delta)\sk
C^{\text{Bernstein-M}}(p,n,\delta) &=&p+ \sqrt{\frac{2p(1-p)}{n} \ell_n(\delta)} + \frac{\ell_n(\delta)}{3n}\\
&\text{where}& \ell_n(\delta) = \eta\log\Big(\frac{\log(n)\log(\eta n)}{\log^2(\eta)\delta}\Big)\text{ with } \eta=1.12\,. \nonumber
\eeqa

\begin{figure}
	\centering
	\includegraphics[width=0.7\linewidth]{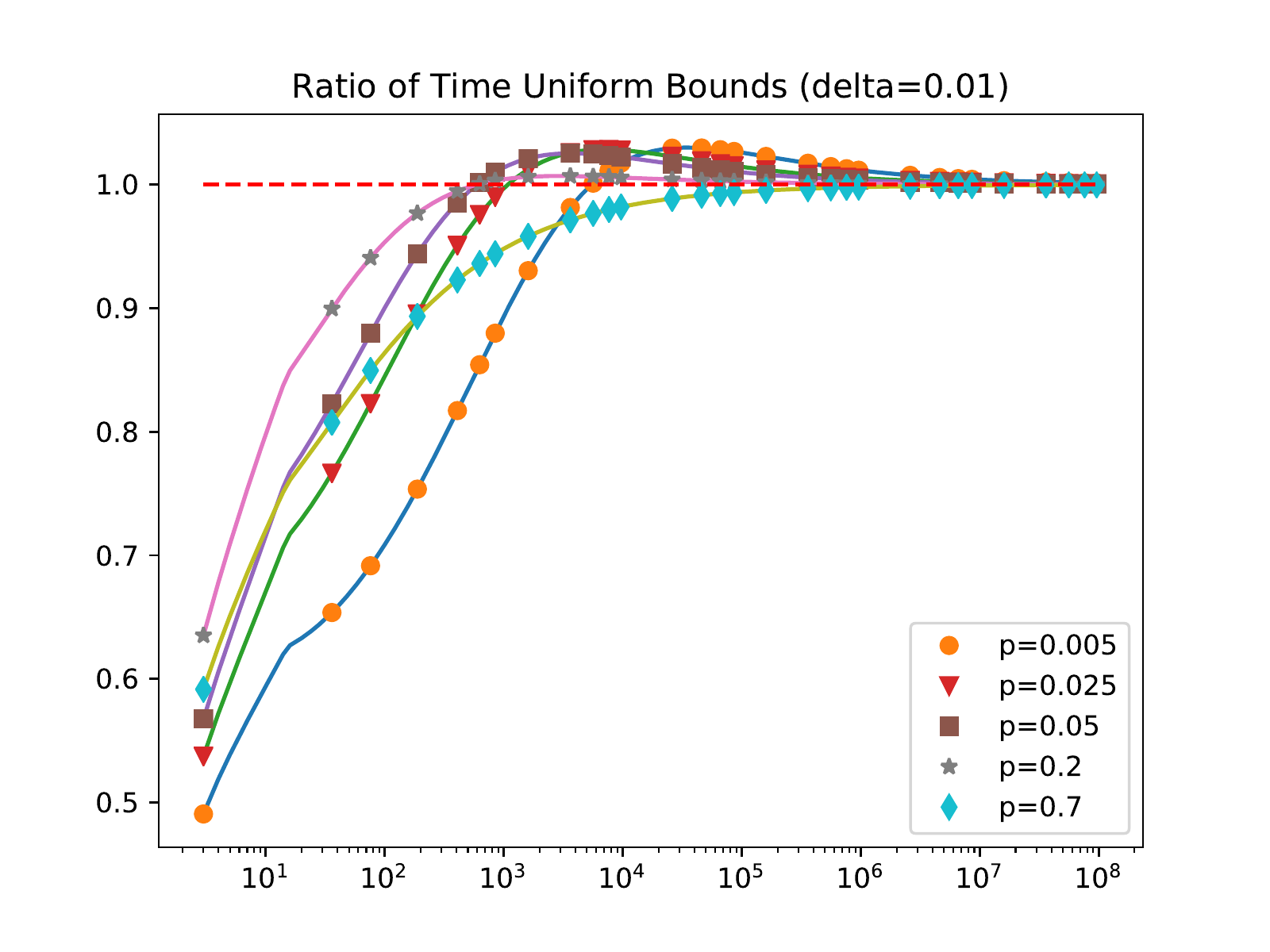}
	\caption{Plot of $n\mapsto r(p,n,\delta)$ for several values of $p$, with $\delta=0.01$. We plot the horizontal line $r(p,n,\delta)=1$ for reference: Above this line, the second Bernstein bound is less tight than the first obe, whereas below this line, the second Bernstein bound is sharper.}
	\label{fig:comparaisonBB}
\end{figure}

Figure~\ref{fig:comparaisonBB} plots the ratio $r(p,n,\delta)=C^{\text{Bernstein-M}}(p,n,\delta)/C^{\text{Bernstein-D}}(p,n,\delta)$
as a function of $n$ for different values of $p$ and for the fixed value of $\delta=0.01$.
This shows the clear advantage of using the considered technique over that of \citep{dann2017unifying}.

\begin{figure}
	\centering
	\includegraphics[width=0.49\linewidth]{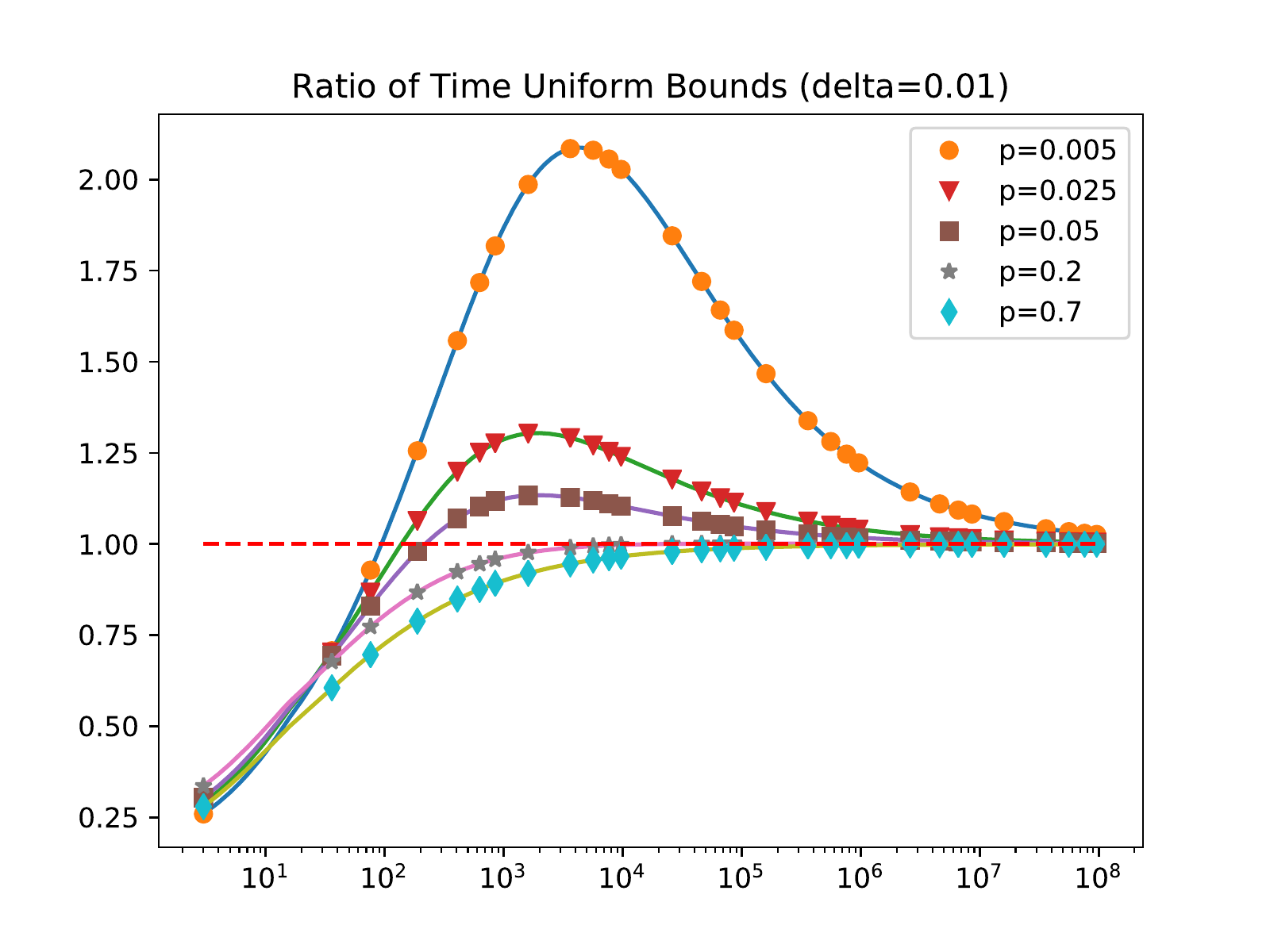}\hfill
	\includegraphics[width=0.49\linewidth]{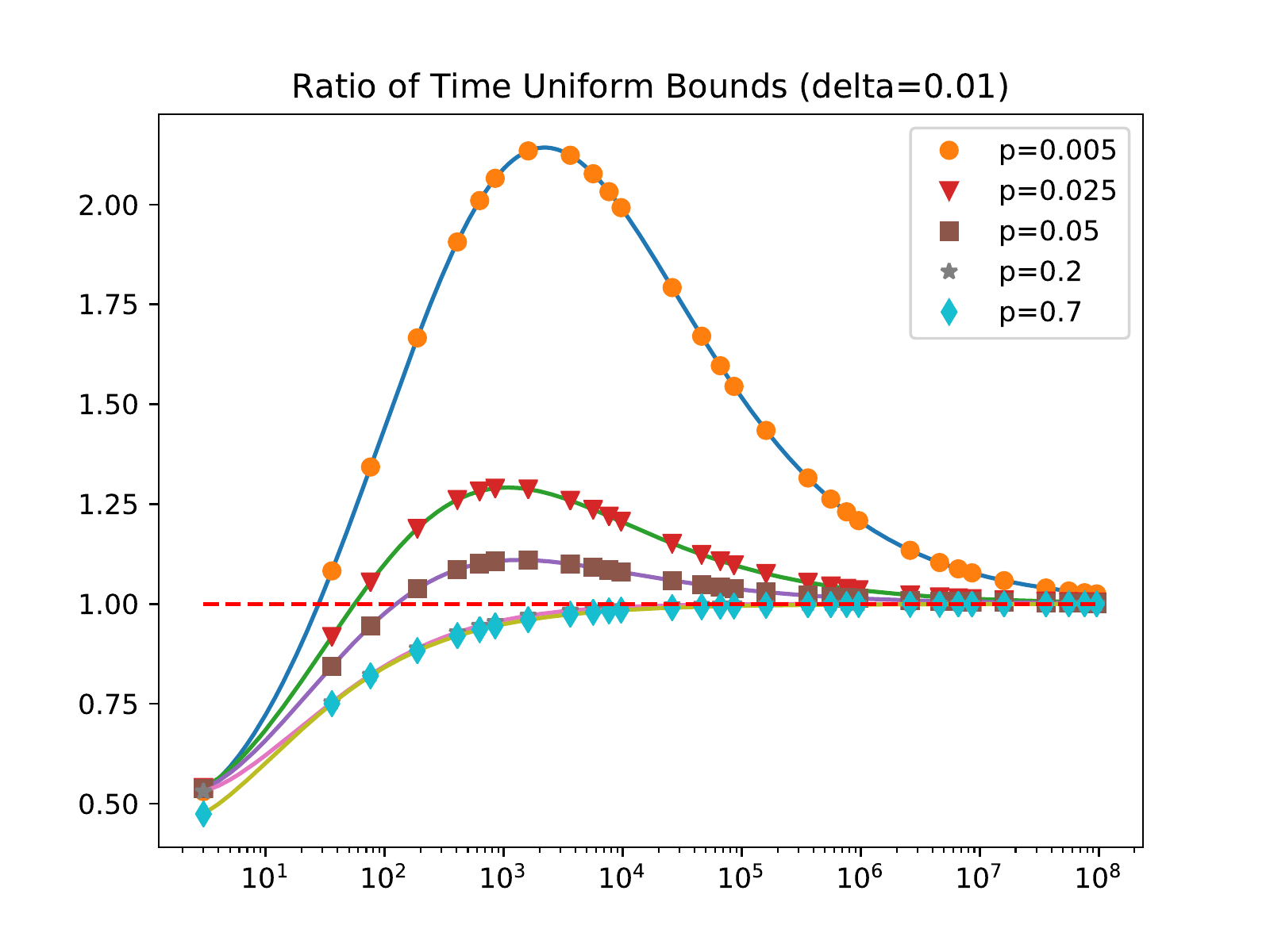}
	\caption{Plot of $n\mapsto r(p,n,\delta)$ for several values of $p$, with $\delta=0.01$. We plot the horizontal line $r(p,n,\delta)=1$ for reference: Above this line, the Gaussian-Laplace bound is looser than the Bernstein bound, while below this line, the Gaussian-Laplace bound is sharper. Left: Using  $C^{\text{Bernstein-D}}$ (the first Bernstein bound). Right: Using $C^{\text{Bernstein-M}}$ (the second Bernstein bound).}
	\label{fig:comparaisongp}
\end{figure}

In order to better understand the benefit of using a sub-Gaussian tail control for Bernoulli, we further introduce the following function
\beqa
\label{eq:CB_gp}
C^{\text{ex-Gaussian-Laplace}}(p,n,\delta) &=& p+\sqrt{\frac{2\underline{g}(p)(1+\frac{1}{n})\log(2\sqrt{n+1}/\delta)}{n} }\,,
\eeqa
and plot in Figure~\ref{fig:comparaisongp} the ratio $r(p,n,\delta)=C^{\text{ex-Gaussian-Laplace}}(p,n,\delta)/C^{\text{e-Bernstein-peeling}}(p,n,\delta)$
as a function of $n$ for different values of $p$ and for the fixed value of $\delta=0.01$. It shows that up to $10^2$ samples (for one state-action pair), (\ref{eq:CB_gp}) is sharper than (\ref{eq:CB_bern_peeling}) for $p>0.005$. Hence, this justifies using (\ref{eq:CB_gp}) in practice.

\section{Extended Value Iteration}\label{app:EVI}

\begin{myproof}{of Lemma \ref{lem:near_opt_supp_selection}}
	By the discussion in Section~\ref{sub:noss} prior to Algorithm~\ref{alg:EVI-NOSS}, we have that

	\beqan
	\Esp_{S\sim p}[f(S)]&=& \sum_{s'\in\widetilde \cS} f(s')p(s') +
	\sum_{s'\notin\widetilde \cS} f(s')p(s')\\
	&\leq& \overline{f}(\widetilde\cS) +
	\sum_{s'\notin\widetilde \cS} f(s')p(s')\\
	&\leq&  \overline{f}(\widetilde\cS) + \min\big(\kappa,\overline{f}(\widetilde\cS),\overline{f}(\cS\setminus\widetilde \cS)\big)\,
	\eeqan
	where the first inequality holds with high probability by Remark~\ref{rem:nearoptsupport}, and the second one is guaranteed by the stopping rule of \texttt{NOSS} (Algorithm~\ref{alg:SupportSelection}).
	Indeed,  \texttt{NOSS} by construction builds a minimal set $\widetilde \cS$ containing the empirical support $\widehat \cS_n$ (plus eventually one point), and satisfies the
	condition $\overline{f}(\cS\setminus\widetilde \cS)<\min\big(\kappa,\overline{f}(\widetilde\cS)\big)$ required to exit the loop.
\end{myproof}

\begin{myproof}{of Lemma~\ref{lem:noEVI}}
	Let us denote by $\star$ an optimal policy. Let ${\bm g}_\star:\cS\to\Real$ denote the constant function equal to  $g_\star$, and ${\bm \kappa_t}$ the constant function equal to  $\kappa_t$.
	Using vector notations, we have on the one hand
	\beqan
	{\bm g}_\star &=& \overline{P}_\star[ \mu_\star + P_\star u_n^+ - u_n^+]\\
	&\leq& \overline{P}_\star[ \mu^+_\star + P_{\star,n}^+ u_n^+ + {\bm \kappa}_t- u_n^+] \text{ w.p. } 1-\delta\\
	&\leq& \overline{P}_\star[ \mu_{\pi_{n+1}^+}^+ + P_{\pi_{n+1}^+,n}^+ u_n^+ - u_n^+] + \overline{P}_\star{\bm \kappa}_t \text{ by  optimality of } \pi_{n+1}^+\\
	&=&\overline{P}_\star[ u_{n+1}^+ - u_n^+] + \overline{P}_\star{\bm \kappa}_t\,.
	\eeqan
	On the other hand, for the MDP computed by \texttt{EVI-NOSS}, it holds
	\beqan
	{\bf g}_{n+1}^+ = \overline{P}_{n+1}^+[\mu_{\pi_{n+1}^+}^+ + P_{n+1}^+u_n^+-u_n^+] = \overline{P}_{n+1}^+[u_{n+1}^+-u_n^+]
	\eeqan
	Hence, combining these two  results, we obtain that with probability higher than $1-\delta$,
	\beqan
	{\bf g}_\star-{\bf g}_{n+1}^+ &\leq& \overline{P}_\star[ u_{n+1}^+ - u_n^+]-\overline{P}_{n+1}^+[u_{n+1}^+-u_n^+]+ \overline{P}_\star{\bf \kappa}_t\\
	&\leq&\bS(u_{n+1}^+ - u_n^+) + \|\overline{P}_\star\|_1\|{\bf \kappa}_t\|_\infty\\
	&\leq& \epsilon + \kappa_t\,.
	\eeqan	
\end{myproof}

\section{Regret Analysis of \UCRLnew: Proof of Lemma \ref{lem:localeffective} and Theorem \ref{thm:regret_UCRLnew}}
In this section, we prove Lemma \ref{lem:localeffective} and Theorem \ref{thm:regret_UCRLnew}. 

\begin{myproof}{of Lemma~\ref{lem:localeffective}}
Recall the definition of the Gini index for pair $(s,a)$: $G_{s,a} := \sum_{x\in \cS} p(x|s,a)(1-p(x|s,a))$. 
Applying Cauchy-Schwarz gives
\als{
	L_{s,a} &= \Big(\sum_{x\in \cK_{s,a}} \sqrt{p(x)(1-p(x))}\Big)^2  \leq K_{s,a}\sum_{x\in \cK_{s,a}}p(x)(1-p(x))  = K_{s,a} G_{s,a}  \, .
}
Furthermore, in view of the concavity of $z\mapsto \sum_{x\in \cS} z(x)(1-z(x))$, the maximal value of $G_{s,a}$ is achieved when $p(x|s,a) = \frac{1}{K_{s,a}}$ for $x\in \cK_{s,a}$. Hence,  $G_{s,a} \leq 1-1/K_{s,a}$. Therefore, $L_{s,a} \leq  K_{s,a} G_{s,a} \leq  K_{s,a} - 1$. 
\end{myproof}

We next prove Theorem \ref{thm:regret_UCRLnew}. Our proof follows similar lines as in the proof of \citep[Theorem~2]{jaksch2010near}. We start with the following time-uniform concentration inequality to control a bounded martingale difference sequence, which follows from Lemma \ref{lem:confintervals}:

\begin{corollary}[Time-uniform Azuma-Hoeffding]\label{coroll:time-uniform-AzumaHoeffding}
	Let $(X_t)_{t\ge 1}$ be a martingale difference sequence such that for all $t$, $X_t\in [a,b]$ almost surely for some $a,b\in \RR$. Then, for all $\delta\in (0,1)$, it holds
	\beqan
	\mathbb P\bigg(\exists T \in\Nat: \sum_{t=1}^T  X_t \geq (b-a)\sqrt{ \tfrac{1}{2}(T+1) \log( \sqrt{T+1}/\delta)}\bigg) \leq \delta\,.
	\eeqan
\end{corollary}

\begin{myproof}{of Theorem \ref{thm:regret_UCRLnew}}
	Let $\delta\in (0,1)$. To simplify notations, we define the short-hand $J_k:=J_{t_k}$ for various random variables that are fixed within a given episode $k$ and omit their dependence on $\delta$ (for example $\cM_{k}:=\cM_{t_k,\delta}$). We let $m(T)$ denote the number of episodes initiated by the algorithm up to time $T$.
	By applying Corollary \ref{coroll:time-uniform-AzumaHoeffding}, we deduce that
	\als{
		\kR(T)&= \sum_{t=1}^T g^\star- \sum_{t=1}^T r_t \leq \sum_{s,a} N_{m(T)}(s,a)(g^\star - \mu(s,a)) + \sqrt{ \tfrac{1}{2}(T+1) \log( \sqrt{T+1}/\delta)}\, ,
	}
	with probability at least $1-\delta$. We have
	\als{
		\sum_{s,a} N_{m(T)}(s,a)(g^\star - \mu(s,a)) &= \sum_{k=1}^{m(T)}  \sum_{s,a}  \sum_{t=t_k}^{t_{k+1}-1}\indic{s_t=s,a_t=a} \big(g^\star - \mu(s,a)\big)\\
		&=
		\sum_{k=1}^{m(T)} \sum_{s,a} \nu_k(s,a) \big(g^\star - \mu(s,a)\big)\,.
	}
	Introducing $\Delta_k := \sum_{s,a}\nu_k(s,a) \big(g^\star - \mu(s,a)\big)$ for $1\leq k\leq m(T)$, we get
	\als{
		\kR(T) \le \sum_{k=1}^{m(T)} \Delta_k + \sqrt{ \tfrac{1}{2}(T+1) \log( \sqrt{T+1}/\delta)}\, ,
	}
	with probability at least $1-\delta$. A given episode $k$ is called \emph{good} if $M \in \cM_{k}$ (that is, the set of plausible MDPs contains the true model), and \emph{bad} otherwise.
	
	\paragraph{Control of the regret due to bad episodes.}
	By Lemma \ref{lem:CI_has_trueMDP}, the set $\cM_{k}$ contains the true MDP with probability higher than $1-\delta$ uniformly for all $T$, and for all episodes $k=1,\ldots,m(T)$. As a consequence, with probability at least $1-\delta$, $
	\sum_{k=1}^{m(T)}\Delta_k\indic{M \notin \cM_{k}} = 0.
	$
	
	\paragraph{Control of the regret due to good episodes.} To upper bound regret in good episodes, we closely follow \citep{jaksch2010near} and decompose the regret to control the transition and reward functions. Consider a good episode $k$ (hence, $M \in \cM_{k}$). By choosing $\pi^+_{k}$ and $\widetilde M_{k}$, using Lemma \ref{lem:noEVI}, we get that
	$$
	g_k := g_{\pi^+_{k}}^{\widetilde M_{k}} \geq g^\star - \frac{1}{\sqrt{t_k}} - \overline\kappa_k \, ,
	$$
	with probability greater than $1-\delta$, where $\overline\kappa_k = \frac{\gamma\bS(u_k) K}{{\max_{s,a}N_k(s,a)^{2/3}}}$. 
	Hence, with probability greater than $1-\delta$,
	\begin{align}
	\Delta_k  &\leq \sum_{s,a} \nu_k(s,a)
	\big(g_k - \mu(s,a)\big) + \sum_{s,a} \nu_k(s,a)\Big(\frac{1}{\sqrt{t_k}}+\overline\kappa_k\Big) 
	\, .
	\label{eq:delta_init}
	\end{align}
	Using the same argument as in the proof of \citep[Theorem~2]{jaksch2010near}, the value function $u_k^{(i)}$ computed by \texttt{EVI-NOSS} at iteration $i$ satisfies: $\max_{s} u_k^{(i)}(s) - \min_{s} u_k^{(i)}(s) \leq D$. The convergence criterion of \texttt{EVI-NOSS} implies
	\begin{equation}
	|u_k^{(i+1)}(s) - u_k^{(i)}(s) - g_k| \leq \frac{1}{\sqrt{t_k}}\,, \qquad  \forall s \in \cS\, .
	\label{eq:puterman_convergence}
	\end{equation}
	Using the Bellman operator on the optimistic MDP, we have:
	\als{
		u_k^{(i+1)}(s) = \widetilde\mu_{k}(s,\pi^+_{k}(s)) + \sum_{s'} \widetilde p_k(s'|s,\pi^+_{k}(s)) u_k^{(i)}(s')\, .
	}
	Substituting this into (\ref{eq:puterman_convergence}) gives
	\begin{equation*}
	\Big|\Big( g_k - \widetilde\mu_{k}(s,\pi^+_{k}(s)) \Big) - \Big(\sum_{s'} \widetilde p_{k}(s'|s,\pi^+_{k}(s))u_k^{(i)}(s') - u_k^{(i)}(s)\Big)\Big|
	\leq \frac{1}{\sqrt{t_k}}\, , \qquad \forall s\in \cS\, .
	\end{equation*}
	Defining
	${\bf g}_k = g_k \mathbf 1$, $\widetilde{\boldsymbol{\mu}}_{k} := \big(\widetilde\mu_{k}(s,\pi^+_k(s))\big)_s$, $\widetilde{\mathbf{P}}_k := \big(\widetilde p_{k}\big(s'|s,\pi^+_k(s)\big)\big)_{s, s'}$ and $\nu_{k} := \big(\nu_k\big(s,\pi^+_k(s)\big)\big)_s$, we can rewrite the above inequality as:
	\begin{equation*}
	\Big|{\bf g}_k - \widetilde{\boldsymbol{\mu}}_{k} - (\widetilde{\mathbf{P}}_k - \mathbf{I}) u_k^{(i)} \Big|
	\leq \frac{1}{\sqrt{t_k}}\mathbf 1\,.
	\end{equation*}
	Combining this with (\ref{eq:delta_init}) yields
	\begin{align}
	\Delta_k&\leq \sum_{s,a} \nu_k(s,a) \big(g_k-\mu(s,a)\big) + \sum_{s,a} \nu_k(s,a)\Big(\frac{1}{\sqrt{t_k}}+\overline\kappa_k\Big) \nonumber\\
	&= \sum_{s,a} \nu_k(s,a) \big( g_k - \widetilde\mu_{k}(s,a) \big) + \sum_{s,a} \nu_k(s,a) \big( \widetilde\mu_{k}(s,a) - \mu(s,a) \big) + \sum_{s,a}
	\nu_k(s,a)\Big(\frac{1}{\sqrt{t_k}} + \overline\kappa_k\Big) \nonumber\\
	&\leq \nu_{k} (\widetilde{\mathbf{P}}_k - \mathbf{I} ) u_k^{(i)} + \sum_{s,a} \nu_k(s,a) \big(\widetilde\mu_{k}(s,a) - \mu(s,a) \big) +  \sum_{s,a} \nu_k(s,a)\Big(\frac{2}{\sqrt{t_k}} + \overline\kappa_k\Big)\, .\nonumber
	\end{align}
	Similarly to \citep{jaksch2010near}, we define $w_k(s) := u_k^{(i)}(s) - \tfrac{1}{2}(\min_s u_k^{(i)}(s) + \max_s u_k^{(i)}(s))$ for all $s\in \cS$. Then, in view of the fact that $\widetilde{\mathbf{P}}_k$ is row-stochastic, we obtain
	\begin{align*}
	\Delta_k&\leq \nu_{k} (\widetilde{\mathbf{P}}_k - \mathbf{I} ) w_k + \sum_{s,a} \nu_k(s,a) \big(\widetilde \mu_{k}(s,a) - \mu(s,a) \big) + \sum_{s,a}  \nu_k(s,a)\Big(\frac{2}{\sqrt{t_k}} + \overline\kappa_k\Big)\, ,
	\end{align*}
	with probability at least $1-\delta$.  
	The second term in the right-hand side can be upper bounded as follows: $M \in \cM_{k}$ implies
	\als{
		\widetilde\mu_{k}(s,a) - \mu(s,a) &\leq 2 b^r_{t, \delta/(3SA(1+S))}(s,a) \\
		&\leq \beta_{N_{k}(s,a)}\big(\tfrac{\delta}{3SA(1+S)}\big) \\
		&= \sqrt{\frac{2}{N_{k}(s,a)}\Big(1\!+\!\frac{1}{N_{k}(s,a)}\Big)\log\Big(3SA(S+1)\sqrt{N_{k}(s,a)\!+\!1}/\delta\Big)} \\
		&\leq \sqrt{\frac{4}{N_k(s,a)}\log\big(6S^2A\sqrt{T+1}/\delta\big)}\, ,
	}
	where we have used $1\leq N_k(s,a) \leq T$ and $S\geq 2$ in the last inequality. Furthermore, using
	$t_k\ge \max_{s,a}N_k(s,a)$ and $\bS(u_k) \leq D$ yields
	\als{
		\sum_{s,a}  \nu_k(s,a)\Big(\frac{2}{\sqrt{t_k}} + \overline\kappa_k\Big) &\le 2\sum_{s,a}  \frac{\nu_k(s,a)}{\sqrt{N_k(s,a)}} + \gamma DK\sum_{s,a} \frac{\nu_k(s,a)}{N_k(s,a)^{2/3}} \, .
	}
	Putting together,  we obtain 
	\begin{align}
	\Delta_k
	&\leq \nu_{k} (\widetilde{\mathbf{P}}_k-\mathbf{I})w_k + \Big(\sqrt{4\log\big(6S^2A\sqrt{T+1}/\delta\big)} + 2\Big)
	\sum_{s,a} \frac{\nu_k(s,a)}{\sqrt{N_{k}(s,a)}} + \gamma DK\sum_{s,a} \frac{\nu_k(s,a)}{N_k(s,a)^{2/3}}\, ,
	\label{eq:main_delta_minmk}\end{align}
	with probability at least $1-\delta$. 
	In what follows, we derive an upper bound on $\nu_{k} (\widetilde{\mathbf{P}}_k-\mathbf{I})w_k$. Similarly to \citep{jaksch2010near}, we consider the following decomposition:
	\als{
		\nu_k(\widetilde{\mathbf{P}}_k - \mathbf{I}) w_k = \underbrace{\nu_k (\widetilde{\mathbf{P}}_k - \mathbf{P}_k) w_k}_{L_1(k)} + \underbrace{\nu_k (\mathbf{P}_k-\mathbf{I})w_k}_{L_2(k)} \, .
	}
	
	The following lemmas provide upper bounds on $L_1(k)$ and $L_2(k)$:
	
	\begin{lemma}\label{lem:L_1_k}
		Consider a good episode $k$. Then,
		\als{
			L_1(k) &\leq  \sqrt{2\ell_{T}\big(\tfrac{\delta}{6S^2A}\big)}\sum_{s,a}\frac{\nu_k(s,a)}{\sqrt{N_k(s,a)}}D_s\sqrt{L_{s,a}} +  4DS\ell_{T}\big(\tfrac{\delta}{6S^2A}\big) \sum_{s,a}\frac{\nu_k(s,a)}{N_k(s,a)}\, .
		}
	\end{lemma}
	
	\begin{lemma}\label{lem:L_2_k}
		For all $T$, it holds with probability at least $1-\delta$,
		\als{
			\sum_{k=1}^{m(T)} L_2(k) \bI\{M \in \cM_{k}\} \leq D \sqrt{2 (T+1) \log(\sqrt{T+1}/\delta)} + DSA\log_2(\tfrac{8T}{SA})\, .
		}
	\end{lemma}
	
	Applying Lemmas \ref{lem:L_1_k} and \ref{lem:L_2_k}, and summing over all good episodes, we obtain the following bound that holds
	with probability higher than $1-2\delta$, uniformly over all $T\in\Nat$:
	\begin{align}
	\sum_{k=1}^{m(T)} \Delta_k &\bI\{M \in \cM_{k}\} \leq
	\sum_{k=1}^{m(T)} L_1(k) + \sum_{k=1}^{m(T)} L_2(k) \nonumber \\
	& + \Big(\sqrt{4\log\big(6S^2A\sqrt{T+1}/\delta\big)} + 2\Big)\sum_{k=1}^{m(T)} \sum_{s,a} \frac{\nu_k(s,a)}{\sqrt{N_{k}(s,a)}} + \gamma DK\sum_{k=1}^{m(T)}\sum_{s,a} \frac{\nu_k(s,a)}{N_k(s,a)^{2/3}} \nonumber \\
	&\leq \sqrt{2\ell_{T}\big(\tfrac{\delta}{6S^2A}\big)}\sum_{s,a}\frac{\nu_k(s,a)}{\sqrt{N_k(s,a)}}D_s\sqrt{L_{s,a}} +  4DS\ell_{T}\big(\tfrac{\delta}{6S^2A}\big) \sum_{s,a}\frac{\nu_k(s,a)}{N_k(s,a)} \sk
	&+ \Big(\sqrt{4\log\big(6S^2A\sqrt{T+1}/\delta\big)} + 2\Big)\sum_{k=1}^{m(T)} \sum_{s,a} \frac{\nu_k(s,a)}{\sqrt{N_{k}(s,a)}} \nonumber \\
	&+ D \sqrt{2 (T+1) \log(\sqrt{T+1}/\delta)} + DSA\log_2(\tfrac{8T}{SA}) + \gamma DK\sum_{k=1}^{m(T)}\sum_{s,a} \frac{\nu_k(s,a)}{N_k(s,a)^{2/3}}\, .
	\label{eq:trans_prob_intermed_minmk}
	\end{align}
	
	To simplify the above bound, we provide the following lemma:
	\begin{lemma}\label{lem:N_k_sequenence_sums}
		We have:
		\als{
			(i)& \qquad \sum_{s,a} \sum_{k=1}^{m(T)} \frac{\nu_k(s,a)}{\sqrt{N_{k}(s,a)}}  \leq
			\big( \sqrt{2} + 1 \big) \sqrt{SAT}\, . \\
			(ii)& \qquad \sum_{s,a} \sum_{k=1}^{m(T)} \frac{\nu_k(s,a)}{\sqrt{N_{k}(s,a)}}D_s\sqrt{L_{s,a}}  \leq
			\big( \sqrt{2} + 1 \big) \sqrt{\sum_{s,a} D_s^2L_{s,a} T}\, . \\
			(iii)& \qquad \sum_{s,a} \sum_{k=1}^{m(T)} \frac{\nu_k(s,a)}{N_{k}(s,a)}  \leq
			2SA\log\big(\tfrac{T}{SA}\big) + SA\, .\\
			(iv)& \qquad \sum_{s,a} \sum_{k=1}^{m(T)} \frac{\nu_k(s,a)}{N_{k}(s,a)^{2/3}}  \leq 6(SA)^{2/3}T^{1/3} + 2SA\, .
		}
	\end{lemma}

	Putting everything together, it holds that with probability at least $1-4\delta$,
	\als{
		\kR(T) &\leq \big( \sqrt{2} + 1 \big)\Big(\sqrt{4\log\big(6S^2A\sqrt{T+1}/\delta\big)} + 2\Big)\sqrt{SAT} + \big(D\sqrt{2} + \sqrt{\tfrac{1}{2}}\big) \sqrt{(T+1)\log(\sqrt{T+1}/\delta)} \\
		&+  \sqrt{2\ell_{T}\big(\tfrac{\delta}{6S^2A}\big)}\big( \sqrt{2} + 1 \big)\sqrt{T\sum_{s,a}D_s^2 L_{s,a}} \\
		&+ 4DS\ell_{T}\big(\tfrac{\delta}{6S^2A}\big) \Big(2SA\log\big(\tfrac{T}{SA}\big) + SA\Big) \\
		&+ 60 DKS^{2/3}A^{2/3} T^{1/3} + DSA\log_2(\tfrac{8T}{SA}) + 20DKSA\, .
	}
	
	Noting that for $S,A\ge 2$, it is easy to verify that for $T\ge 3$, 
	$\ell_{T}\big(\tfrac{\delta}{6S^2A}\big) \leq 2\log\big(6S^2A\sqrt{T+1}/\delta\big)$. Hence, after simplification we obtain that for all $T\ge 3$, with probability at least $1-4\delta$,  
	\als{
		\kR(T) &\leq \bigg(5\sqrt{\sum\nolimits_{s,a} D_s^2 L_{s,a}} + 10\sqrt{SA} + 2D \bigg)\sqrt{T\log\Big(\tfrac{6S^2A\sqrt{T+1}}{\delta}\Big)}
		+ 60 DKS^{2/3}A^{2/3} T^{1/3} + \cO\Big(DS^2A\log^2\big(\tfrac{T}{\delta}\big) \Big)\, .
	}
	
	Finally we remark that 
	\als{
		5\sqrt{\sum\nolimits_{s,a} D_s^2 L_{s,a}} + 10\sqrt{SA} \leq 20\sqrt{SA \!+\! \sum\nolimits_{s,a} D_s^2L_{s,a} } \leq 20\sqrt{2\sum\nolimits_{s,a} \big(D_s^2L_{s,a}\lor 1\big)}\, ,
	}
	so that $
	\kR(T) = \cO\Big(\Big[\sqrt{\sum\nolimits_{s,a} \big(D_s^2L_{s,a}\lor 1\big)} + D\Big]\sqrt{T\log(\sqrt{T}/\delta)}\Big).
	$
\end{myproof}

\subsection{Proof of Technical Lemmas}

\begin{myproof}{of Lemma \ref{lem:L_1_k}}
	To derive an upper bound on $L_1(k)$, first notice that
	\als{
		L_1(k) &= \sum_{s,x}\nu_k(s,\pi_k^+(s)) \Big(\widetilde p_{k}(x|s,\pi_k^+(s)) - p(x|s,\pi_k^+(s))\Big) w_k(x) \sk
		&\le \sum_{s,a}\nu_k(s,a) \sum_x \Big(\widetilde p_{k}(x|s,a) - p(x|s,a)\Big) w_k(x) \, .
	}
	
	Fix $s$ and $a$, and introduce short-hands $N_k:=N_k(s,a)$, $\widetilde p_{k}:=\widetilde p_{k}(\cdot|s,a)$, $\widehat p_{k}:=\widehat p_{k}(\cdot|s,a)$, and $p:=p(\cdot|s,a)$. We have
	\als{
		\sum_{x}\Big(\widetilde p_{k}(x|s,a)&- p_k(x|s,a)\Big) w_k(x)= \sum_{x} (\widetilde p_{k}(x) - p(x)) w_k(x) \sk
		&\leq \underbrace{\sum_x  |\widehat p_{k}(x) - p(x)||w_k(x)|}_{F_1} + \underbrace{\sum_{x} |\widetilde p_{k}(x)- \widehat p_{k}(x)||w_k(x)|}_{F_2} \, .
	}
	
	To upper bound $F_1$, we first show that $\max_{x\in \supp(\widetilde p_k(\cdot|s,a))}|w_k(x)| \leq \tfrac{D_s}{2}$. To show this, we note that similarly to \cite{jaksch2010near}, we can combine all MDPs in $\cM_k$ to form a single MDP $\widetilde\cM_k$ with continuous action space $\cA'$. In this extended MDP, in each state $s\in \cS$, and for each $a\in \cA$, there is an action in $\cA'$ with mean $\widetilde\mu(s,a)$ and transition probability $\widetilde p(\cdot|s,a)$ satisfying (\ref{eq:UCRLnew_CB}). Similarly to \cite{jaksch2010near}, we note  that $u^{(i)}_k(s)$ amounts to the total expected $i$-step reward of an optimal non-stationary $i$-step policy starting in state $s$ on the MDP $\widetilde\cM_k$ with the extended action set. The local diameter of state $s$ of this extended MDP is at most $D_s$, since by assumption $k$ is a good episode and hence $\cM_k$ contains the true MDP $M$, and therefore, the actions of the true MDP are contained in the continuous action set of $\widetilde\cM_k$. Now, if there were states $s_1,s_2\in \cup_{a} \supp(\widetilde p_k(\cdot|s,a))$ with $u^{(i)}_k(s_1) - u^{(i)}_k(s_2) > D_s$, then an improved value
for $u^{(i)}_k(s_1)$ could be achieved by the following non-stationary policy: First follow a policy that
moves from $s_1$ to $s_2$ most quickly, which takes at most $D_s$ steps on average. Then follow the optimal
$i$-step policy for $s_2$. We thus have $u^{(i)}_k(s_1) \geq u^{(i)}_k(s_2)-D_s$, since at most $D_s$ of the $i$ rewards of the policy for $s_2$ are missed. This is a contradiction, and so the claim follows. 
	
	To upper bound $F_1$, 	noting that $k$ is a good episode yields:
	\als{
		F_1  &\leq \sqrt{\frac{2\ell_{N_k}}{N_k}}\sum_x \sqrt{p(x)(1-p(x))}|w_k(x)| + \frac{S\ell_{N_k}}{3N_k}\|w_k\|_\infty\\
		&\leq \max_{x\in \cK_{s,a}}|w_k(x)| \sqrt{\frac{2\ell_{N_k}}{N_k}}\sum_{x} \sqrt{p(x)(1-p(x))} + \frac{DS\ell_{N_k}}{6N_k} \\
		&\leq D_s \sqrt{\frac{\ell_{N_k}}{2 N_k}}\sum_{x} \sqrt{p(x)(1-p(x))} + \frac{DS\ell_{N_k}}{6N_k} \\
		&= D_s \sqrt{\frac{\ell_{N_k}}{2 N_k}L_{s,a}} + \frac{DS\ell_{N_k}}{6N_k}
		\,,
	}
	where we have used that $\|w_k\|_\infty\leq \tfrac{D}{2}$ and $\max_{x\in \cK_{s,a}}|w_k(x)| \leq \tfrac{D_s}{2}$.

	To upper bound $F_2$, we will need the following lemma:
	\begin{lemma}
		\label{lem:SqrtVar_pqBern}
		Consider $x$ and $y$ satisfying $|x - y| \leq \sqrt{2y(1-y)\zeta} + \zeta/3$. Then,
		\begin{align*}
		\sqrt{y(1 - y)} &\leq \sqrt{x(1- x)} + 2.4\sqrt{\zeta} \, .
		\end{align*}
	\end{lemma}
	
	Applying Lemma \ref{lem:SqrtVar_pqBern} twice and using the relation $\max_{x\in \supp(\widetilde p_k(\cdot|s,a))}|w_k(x)| \leq \tfrac{D_s}{2}$ yield:
	\als{
		F_2  &\leq \sqrt{\frac{2\ell_{N_k}}{N_k}}\sum_x \sqrt{\widetilde p_{k}(x)(1-\widetilde p_{k}(x))}|w_k(x)| + \frac{DS\ell_{N_k}}{6N_k} \\
		 &\leq D_s\sqrt{\frac{\ell_{N_k}}{2N_k}}\sum_x \sqrt{\widetilde p_{k}(x)(1-\widetilde p_{k}(x))} + \frac{DS\ell_{N_k}}{6N_k} \\
		&\leq D_s\sqrt{\frac{\ell_{N_k}}{2N_k}}\sum_x \sqrt{\widehat p_{k}(x)(1-\widehat p_{k}(x))} + 2.4\sqrt{2}\frac{DS\ell_{N_k}}{N_k} + \frac{DS\ell_{N_k}}{6N_k} \\
		&\leq D_s\sqrt{\frac{\ell_{N_k}}{2N_k}}\sum_x \sqrt{p(x)(1- p(x))} + \frac{3.6DS\ell_{N_k}}{N_k} \,.
	}
	Combininig the bounds on $F_1$ and $F_2$, and noting that 
	\als{
		\ell_{N_k(s,a)}\big(\tfrac{\delta}{3(1+S)SA}\big) \leq \ell_{N_k(s,a)}\big(\tfrac{\delta}{6S^2A}\big) \leq \ell_{T}\big(\tfrac{\delta}{6S^2A}\big)
	}
	complete the proof. 
\end{myproof}

\begin{myproof}{of Lemma \ref{lem:SqrtVar_pqBern}}
	By Taylor's expansion, we have
	\als{
		y(1-y) &= x (1- x) + (1 - 2x)(y - x) -(y - x)^2 \\
		&= x(1- x) + (1 - x - y)(y - x) \\
		&\leq
		x (1-x) + |1 - x - y |\left(\sqrt{2y (1-y)\zeta} + \tfrac{1}{3}\zeta\right) \sk
		&\leq
		x (1-x) + \sqrt{2 y(1- y)\zeta} + \tfrac{1}{3}\zeta
		\, .
	}
	Using the fact that $a\le b\sqrt{a}+c$ implies $a\le b^2 + b\sqrt{c} + c$ for nonnegative numbers $a,b$, and $c$, we get
	\begin{align}
	y(1-y)
	&\leq
	x (1-x)  + \tfrac{1}{3}\zeta + \sqrt{2\zeta\left(x (1-x)  + \tfrac{1}{3}\zeta\right)} + 2\zeta \sk
	&\leq
	x (1-x)  + \sqrt{2\zeta x (1-x)} + 3.15\zeta  \sk
	&= \left(\sqrt{x (1-x)}  + \sqrt{\tfrac{1}{2}\zeta}\right)^2 + 2.65\zeta\, ,
	\end{align}
	where we have used $\sqrt{a+b}\le \sqrt{a} + \sqrt{b}$ valid for all $a,b\ge 0$. Taking square-root from both sides and using the latter inequality give the desired result:
	\begin{align*}
	\sqrt{y(1 - y)} &\leq
	\sqrt{x(1- x)} + \sqrt{\tfrac{1}{2}\zeta} + \sqrt{2.65\zeta} \leq \sqrt{x(1- x)} + 2.4\sqrt{\zeta} \, .
	\end{align*}
\end{myproof}

\begin{myproof}{of Lemma \ref{lem:L_2_k}}
	Similarly to the proof of \citep[Theorem~2]{jaksch2010near}, we define the sequence $(X_t)_{t\geq 1}$ with $X_t := (p(\cdot|s_t,a_t) - \mathbf e_{s_{t+1}})w_{k(t)}\bI\{M \in \cM_{k(t)}\}$, for all $t$, where $k(t)$ denotes the episode containing time step $t$. For any $k$ with $M \in \cM_{k}$, we have that:
	\als{
		L_2(k) &= \nu_k (\mathbf{P}_k-\mathbf{I})w_k  = \sum_{t=t_k}^{t_{k+1} -1} (p(\cdot|s_t,a_t) - \mathbf{e}_{s_t}) w_k \\
		&= \sum_{t=t_k}^{t_{k+1} -1}
		\Big(p(\cdot|s_t,a_t) - \textbf{e}_{s_{t+1}} + \textbf{e}_{s_{t+1}} - \textbf{e}_{s_t} \Big) w_k 	= \sum_{t=t_k}^{t_{k+1} -1} X_t + w_k(s_{t+1}) - w_k(s_t) \leq \sum_{t=t_k}^{t_{k+1} -1} X_t + D \, ,
	}
	so that $\sum_{k=1}^{m(T)} L_2(k) \leq \sum_{t=1}^T
	X_t +m(T)D$. Using $\|w_k\|_{\infty} = \frac{D}{2}$ and applying the H\"{o}lder inequality give
	$$|X_t| \leq \|p(\cdot|s_t,a_t) - \mathbf e_{s_{t+1}}\|_1\frac{D}{2} \leq \Big(\|p(\cdot|s_t,a_t)\|_1 + \|\mathbf e_{s_{t+1}}\|_1\Big)\frac{D}{2}  = D\,.$$
	So, $X_t$ is bounded by $D$, and also $\mathbb{E}[X_t|s_1, a_1, \dots, s_t, a_t] = 0$, so that $(X_t)_{t\ge 1}$ is martingale difference sequence. Therefore, by  Corollary  \ref{coroll:time-uniform-AzumaHoeffding}, we get:
	\beqan
	\mathbb P\bigg(\exists  T: \sum_{t=1}^T X_t  \geq D \sqrt{2 (T+1) \log(\sqrt{T+1}/\delta)}\bigg)\leq \delta\,,
	\eeqan
thus concluding the proof. 
\end{myproof}


\subsection{Proof of Supporting Lemmas}

\begin{myproof}{of Lemma \ref{lem:N_k_sequenence_sums}}
	Inequalities (i)-(iii) easily follow from Lemma \ref{lem:sequence_sums}, which is stated at the end of this proof, and using Jensen's inequality. Next we prove the inequality (iv).
	
	Given $t\ge 1$, let $k(t)$ denote the episode containing time step $t$. Following similar steps as in the proof of \citep[Lemma~5]{ouyang2017learning}, we have
	\als{
		\sum_{s,a}\sum_{k=1}^{m(T)} \frac{\nu_k(s,a)}{N_k(s,a)^{2/3}} &= \sum_{s,a}\sum_{t=1}^T \frac{\indic{(s_t,a_t)=(s,a)}}{N_{k(t)}(s,a)^{2/3}} \\
		&\leq 2\sum_{s,a}\sum_{t=1}^T \frac{\indic{(s_t,a_t)=(s,a)}}{N_t(s,a)^{2/3}} \\
		&= 2\sum_{s,a}\left(\indic{N_{m(T)}(s,a)\geq 1} + \sum_{j=1}^{N_{m(T)}(s,a)} j^{-2/3} \right) \\
		&\leq 2SA + 6\sum_{s,a} N_{m(T)}(s,a)^{1/3} \\
		&\leq 2SA + 6SA\left(\sum_{s,a}\frac{N_{m(T)}(s,a)}{SA}\right)^{1/3} \\
		&= 2SA + 6S^{2/3}A^{2/3}T^{1/3} \, ,
	}
	where we have used that for any $L\geq 1$, $\sum_{j=1}^L j^{-2/3} \leq 1+ \int_{1}^L z^{-2/3}\mathrm{d}z \leq 3L^{1/3}$, and where the last step follows from Jensen's inequality.
\end{myproof}

\begin{lemma}[{\citep[Lemma~19]{jaksch2010near},\citep[Lemma~24]{talebi2018variance}}]\label{lem:sequence_sums}
	For any sequence of numbers $z_1, z_2, \dots, z_n$ with $0 \leq z_k \leq Z_{k-1} := \max\{1, \sum_{i=1}^{k-1}z_i\}$, it holds
	\als{
		(i)& \qquad \sum_{k=1}^n 	\frac{z_k}{\sqrt{Z_{k-1}}} 	\leq 	\big(\sqrt{2} + 1\big) 	\sqrt{Z_n}\, . \\
		(ii)& \qquad \sum_{k=1}^n \frac{z_k}{Z_{k-1}} \leq 2\log(Z_n) + 1\, .\\
	}
\end{lemma}

\section{Further Details for Experiments}\label{app:xpsdetails}
\label{sec:details}

\paragraph{Tie-breaking rule to compute optimistic policies.}
All the  considered algorithms (\UCRL, \KLUCRL, \UCRLB, \UCRLnew) resort to a form of \EVI\ internal procedure, that computes
	at each iteration $n$ a policy $\pi_n^+$ maximizing the current optimistic value $u_n^+$ (see Algorithm~\ref{alg:EVI}).
	In practice, several policies may satisfy this, hence a tie-breaking rule is required. For fairness, we used the same tie-breaking rule for all algorithms. It consists, for a state $s$, to break ties by defining the policy to choose an action uniformly randomly amongst $\Argmin_{a\in\cA}N_k(s,a)$. Such breaking rules aim to stabilize the algorithm.

\paragraph{Atypical sequences.} 
The concentration inequalities we have employed for \UCRLnew\ are mostly tight. 
Unfortunately, concentration inequalities are also known to be loose in the specific case of atypical sequences of observations.
Namely, the specific situation when $n=N_t(s,a)>1$ and all observed samples from $(s,a)$ equal $s_0$, 
corresponds to observing a sequence of $n$ ones from a Bernoulli distribution with parameter $\theta=p(s_0|s,a)$.
Note that for $n$ i.i.d.~observations, this event should be of probability $\theta^n$.
In such a situation where $\widehat p_t(s_0|s,a)=1$, all concentration inequalities yield conservative lower bounds on $p(s_0|s,a)$.
We replace these lower bounds with $(1/2)^n$ for this very specific situation.

\paragraph{Extended Value Iterations with lazy support updates}
The \texttt{EVI-NOSS} procedure proceeds in steps, first computing an optimistic support,
then updating $u$ and $\pi$ using the Bellman optimal operator at every single step.
In order to reduce computation, we use a lazy implementation that keeps updating $u$ and $\pi$ at each step but updates the support only once every $L$-steps. This also tends to reduce the number of steps before convergence in practice. In our experiments, we chose $L=5$.

\paragraph{Code release}
The full code implementation is made publicly available as an article companion following this link:  
\texttt{https://gitlab.inria.fr/omaillar/average-reward-reinforcement-learning}. It is coded in Python 3, and is designed to be compatible with OpenAI gym discrete environments.

\section{Numerical Experiments with \PSRL}
In this section, we provide further numerical comparison with a version of the \PSRL\ algorithm \citep{osband2013more} for average-reward RL. \PSRL\ is a popular algorithm originally designed and analysed for episodic RL, and to the best of our knowledge, its (frequentist) regret guarantees in the context of average-reward regret minimization are still unclear.
In this section we will show that \PSRL\ can be a competitive strategy in several environments but also, unfortunately, completely fails in some others. We believe might provide pointers to the lack of theoretical guarantees for this strategy and suggests further modifications could help obtain the best of both worlds (\UCRLnew\ and \PSRL). 

We study a variant of \PSRL, which maintains for each state-action pair $(s,a)$ a Dirichlet distribution to model the transition distribution $p(\cdot|s,a)$, and a Beta distribution to model the reward distribution $\nu(s,a)$.
The Beta distribution is classically used as a prior to model Bernoulli distributions. Here, we only know that the rewards are supported on $[0,1]$, but we can use the popular \emph{Bernoullization} trick. That is, using $n$ rewards sampled from $\nu(s,a)$, we use a Beta distribution $\mathrm{Beta}(S+\alpha,F+\alpha)$, where $S$ stands for the pseudo-success-counts equal to the sum of the $n$ rewards, and $F$ denotes the pseudo-failure-counts equal to $n-S$. 
The Dirichlet distribution is initialized with uniform weights equal to $\alpha$, and we use $\alpha=1$ for both Dirichlet and Beta initial parameters.

We report in the next figures the results of \PSRL\ against \UCRLnew\ and some algorithms that enjoy controlled (frequentist) regret guarantees -- In the figures, we referred to this variant as \PSRLAVR. 

\begin{figure}[htbp]
	\begin{minipage}[r]{0.5\textwidth}
		\begin{center}
			\includegraphics[width=0.9\linewidth, trim = {0 0 0 10mm}, clip]{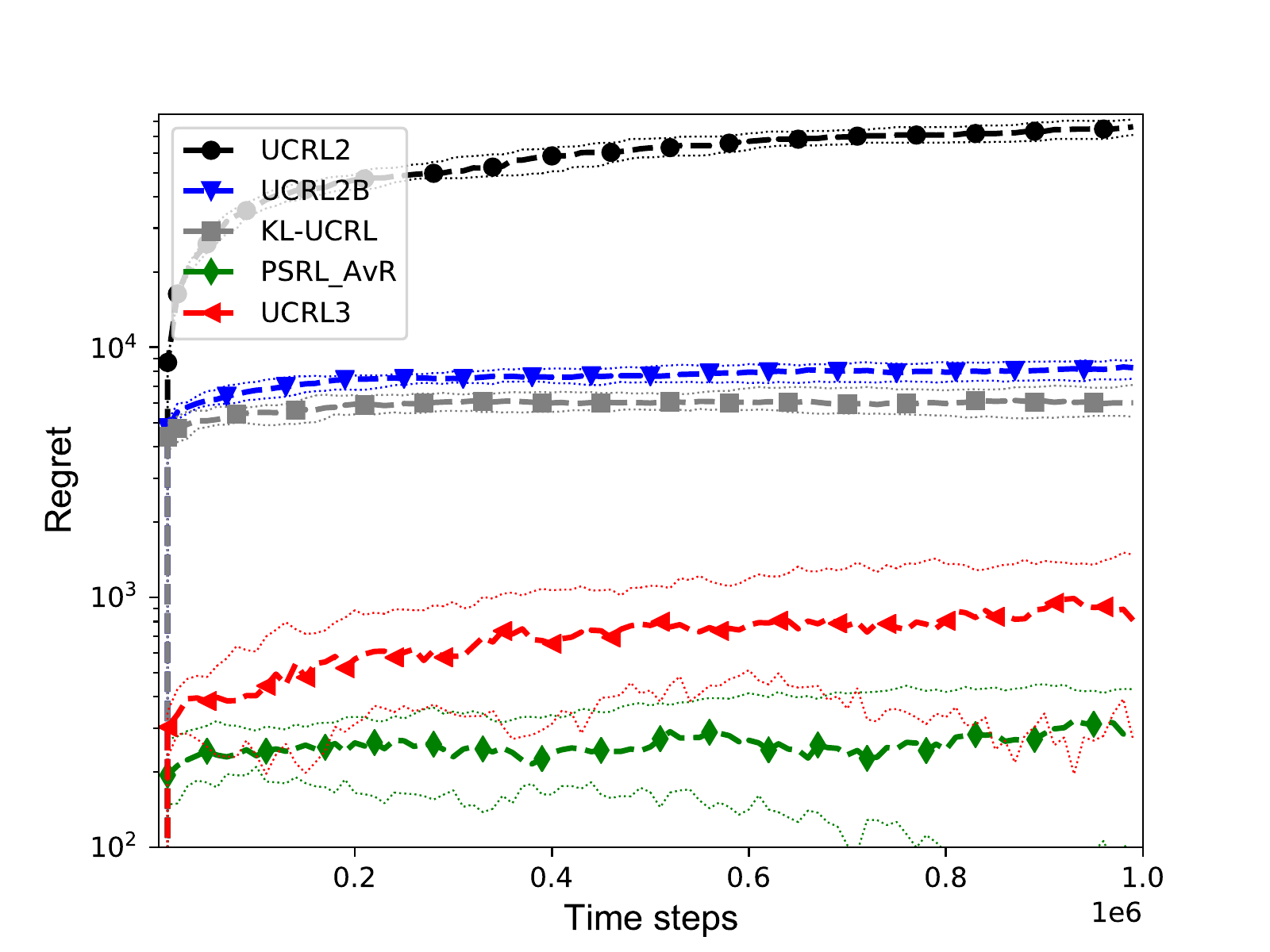}
		\end{center} 
	\end{minipage}\hfill
\begin{minipage}[r]{0.5\textwidth}
	\begin{center}
		\includegraphics[width=0.9\linewidth, trim = {0 0 0 10mm}, clip]{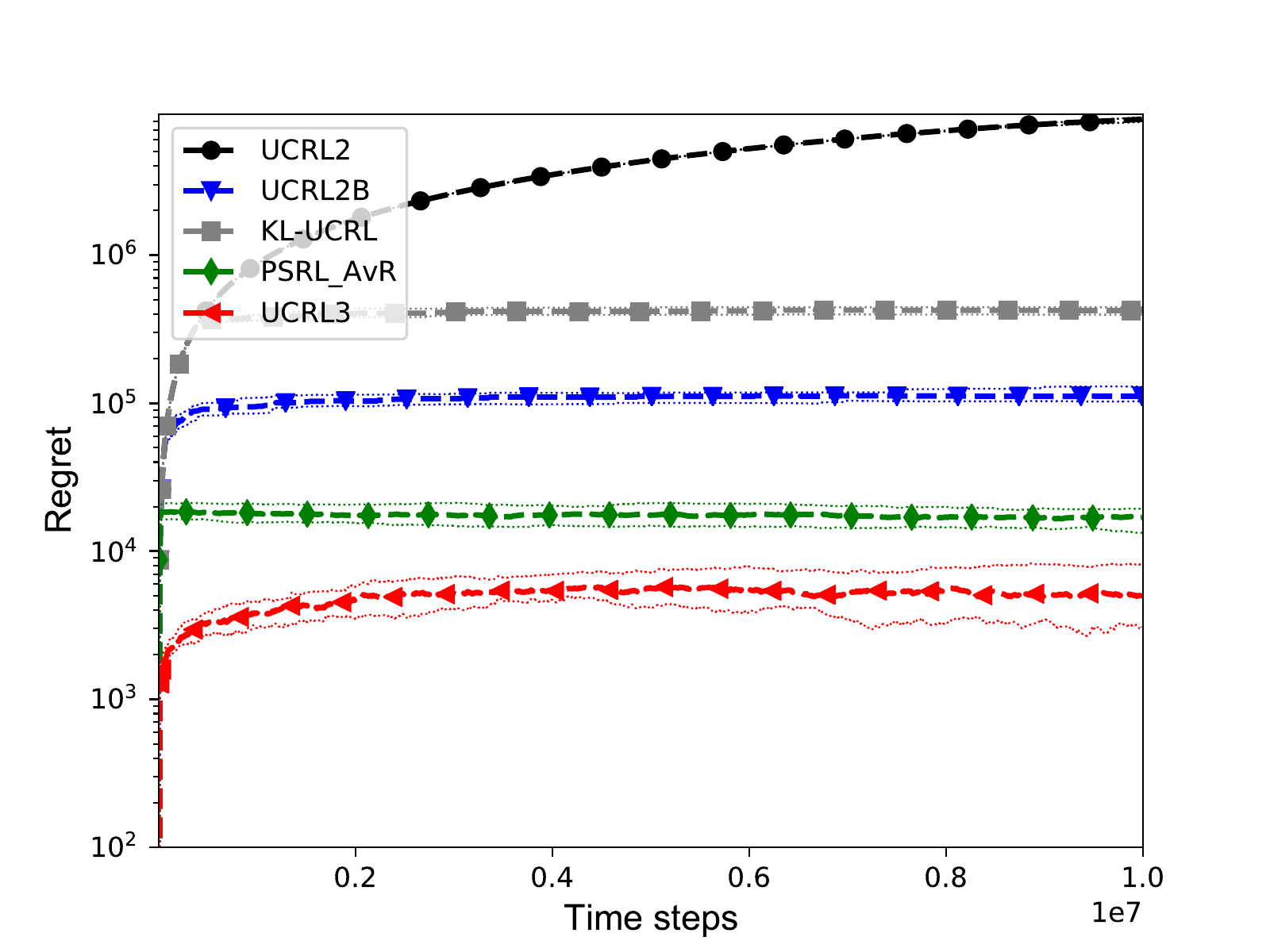}
	\end{center}
\end{minipage}\\	
\vspace{-3mm}
			\caption{Regret for the 6-state (left) and 25-state (right) \textit{RiverSwim} environments}   	
			\vspace{-5mm}
			\label{fig:PSRLriverswim}
\end{figure}

\begin{figure}[htbp]
	\begin{minipage}[r]{0.5\textwidth}
		\begin{center}
			\includegraphics[width=0.9\linewidth, trim = {0 0 0 10mm}, clip]{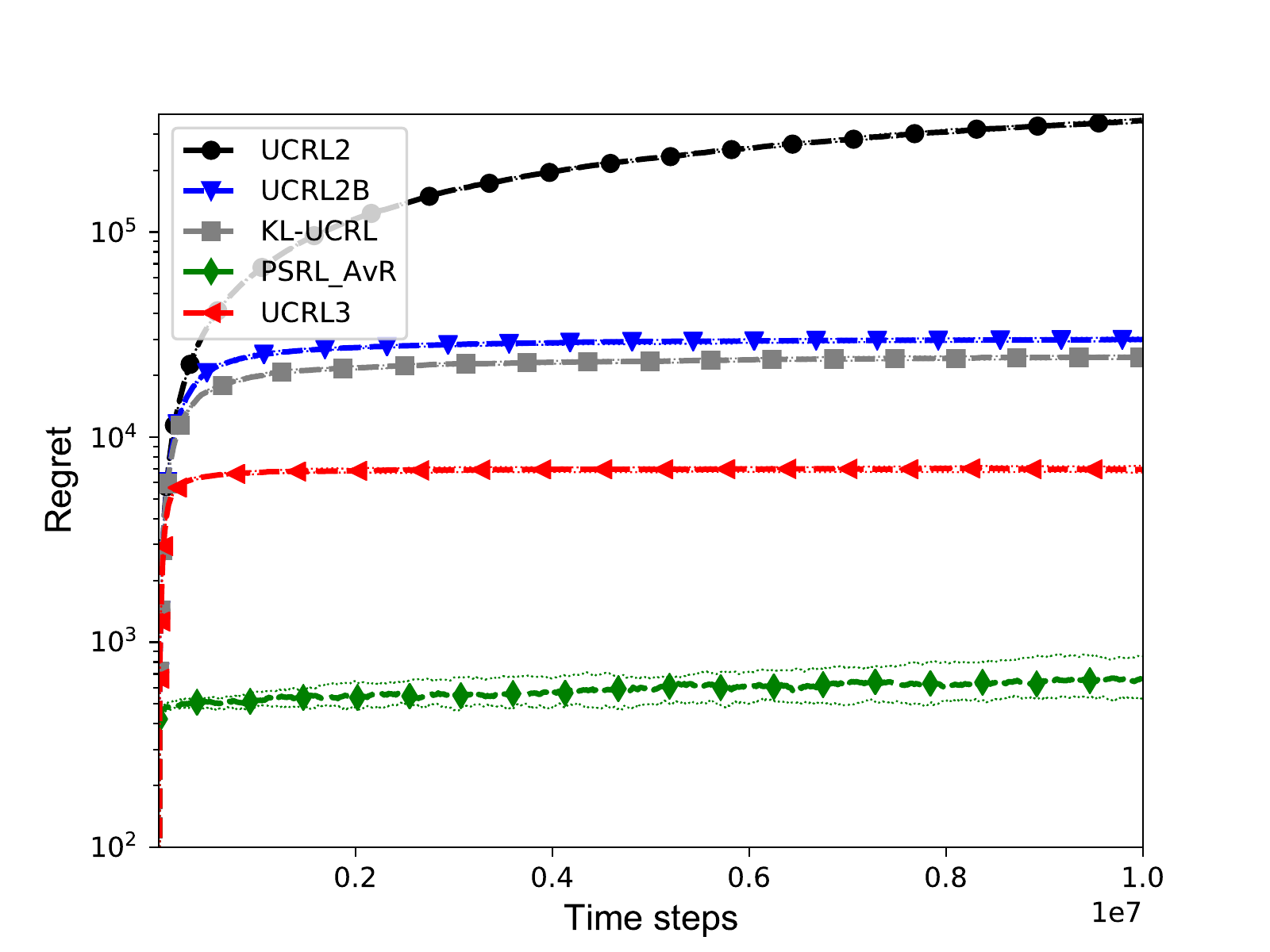}\\
		\end{center}
	\end{minipage}\hfill
\begin{minipage}[r]{0.5\textwidth}
	\begin{center}
		\includegraphics[width=0.9\linewidth, trim = {0 0 0 10mm}, clip]{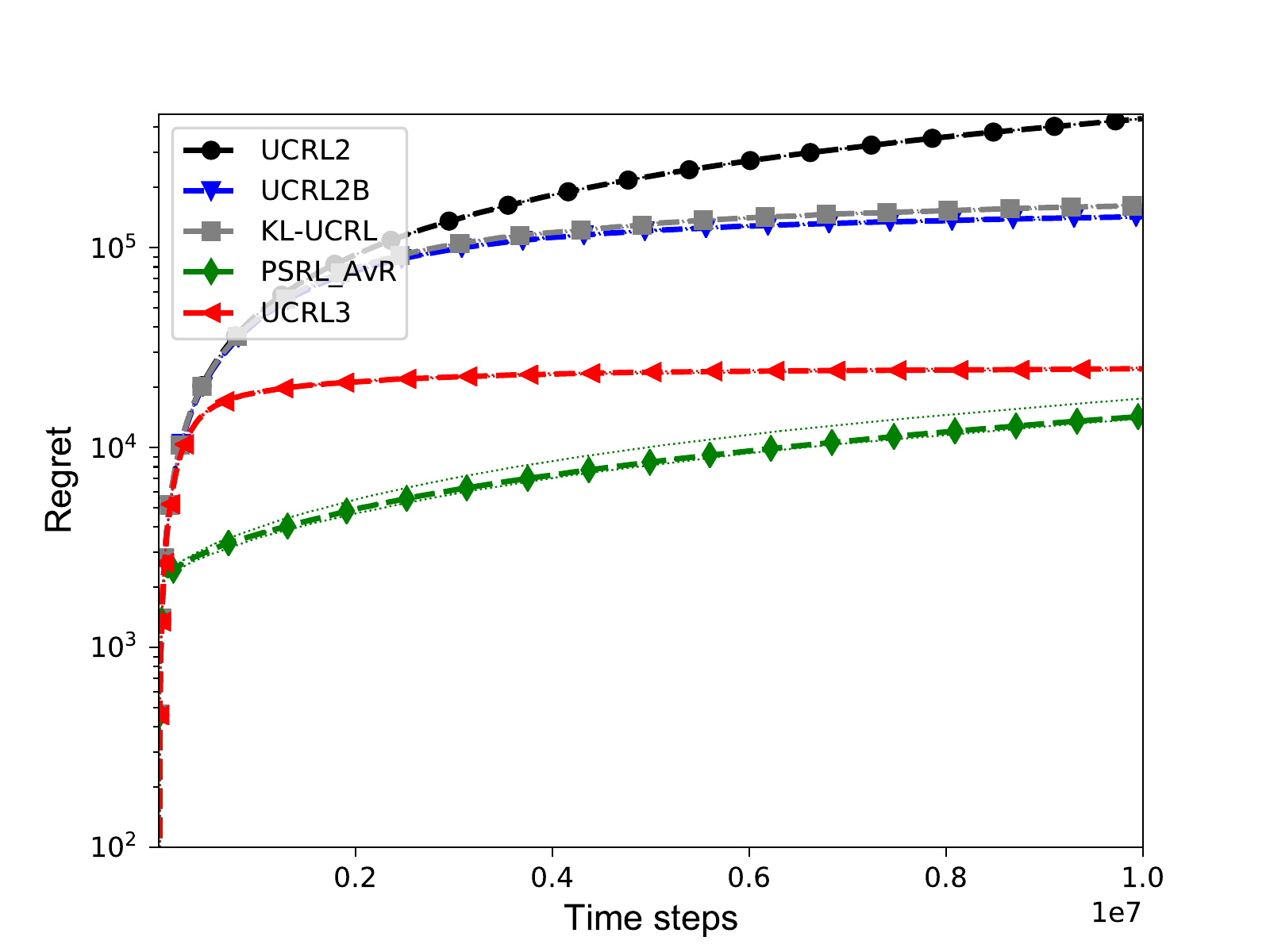}\\
	\end{center}
\end{minipage}\\	
			\vspace{-3mm}
			\caption{Regret for the 2-room (left) and 4-room (right) goal-state environments}
			\vspace{-5mm}
			\label{fig:PSRLrooms}
\end{figure}

\begin{figure}[htbp]
	\begin{minipage}[r]{0.5\textwidth}
		\begin{center}
			\includegraphics[width=0.9\linewidth, trim = {0 0 0 10mm}, clip]{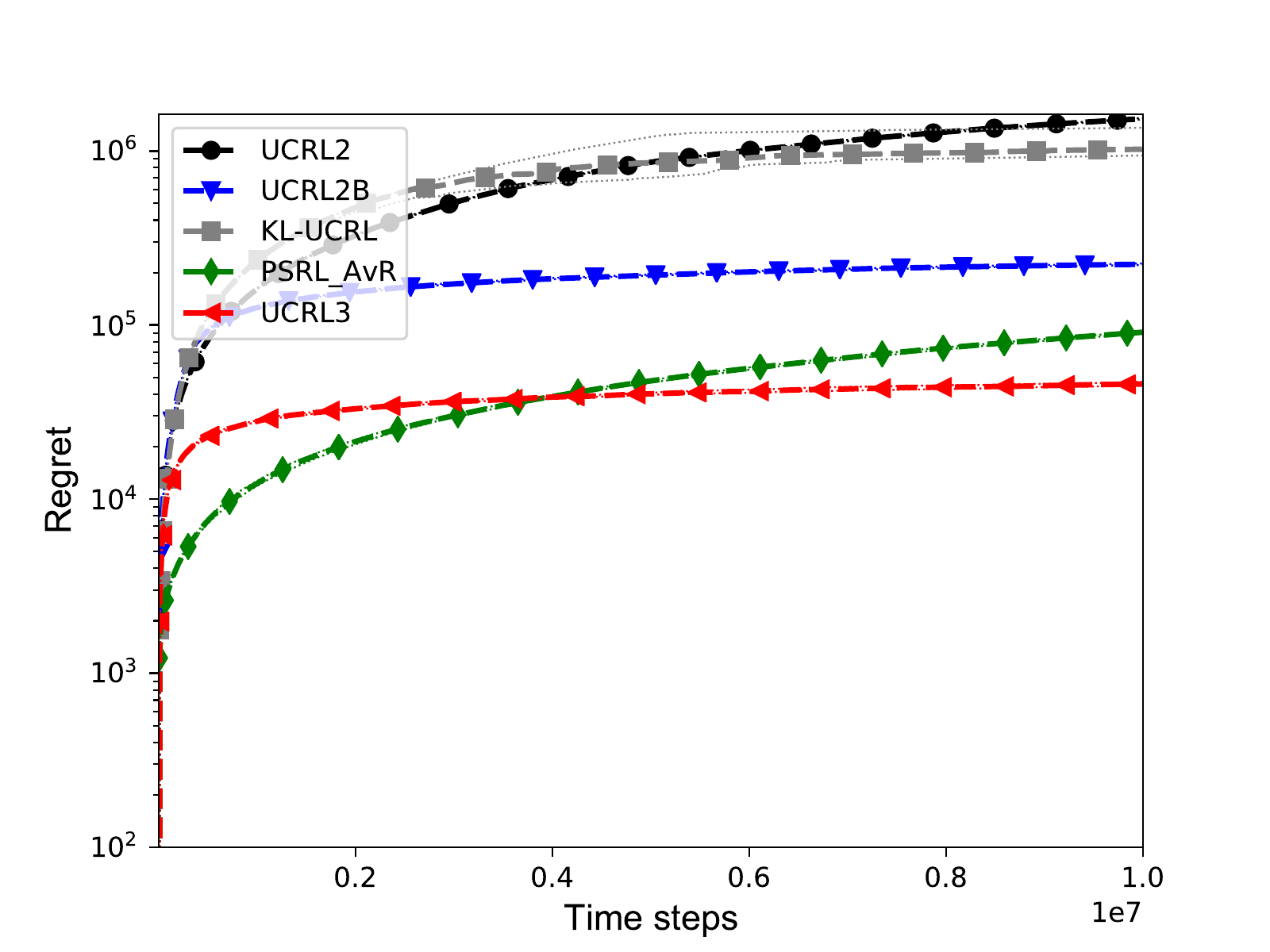}\\
		\end{center}
	\end{minipage}\hfill
	\begin{minipage}[r]{0.5\textwidth}
	\begin{center}
		\includegraphics[width=0.9\linewidth, trim = {0 0 0 10mm}, clip]{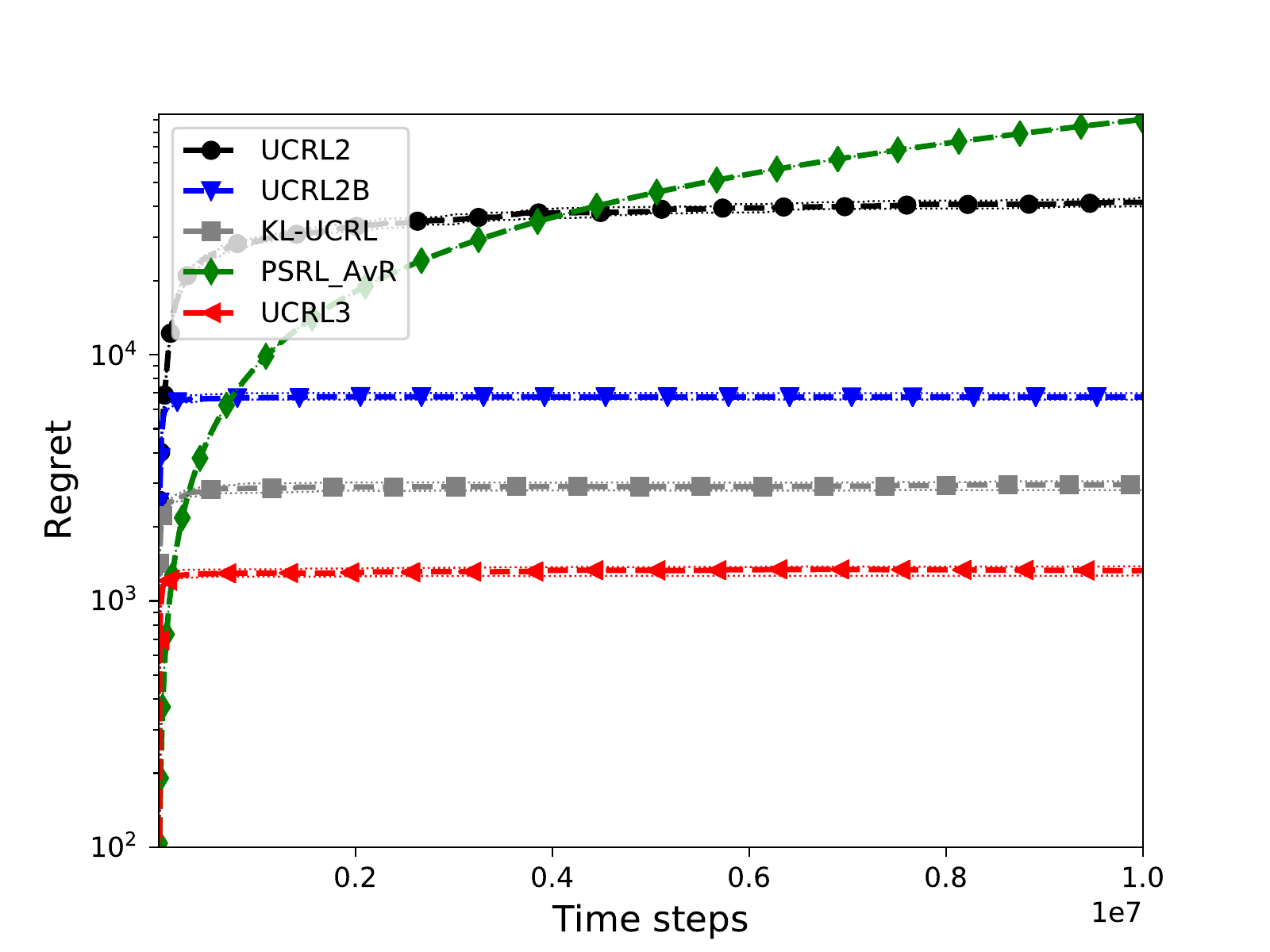}\\
	\end{center}
\end{minipage}\\
			\vspace{-3mm}
			\caption{Regret in a $100$-state randomly generated MDP with sparse rewards (left), and a $10$-state  randomly generated MDP with rich rewards (right)}
			\vspace{-5mm}
			\label{fig:PSRLrandomMDP}
\end{figure}

\begin{figure}[hbtp]
	\begin{center}
		\includegraphics[scale=.27, trim = {0cm 0.cm 0 0.cm},clip]{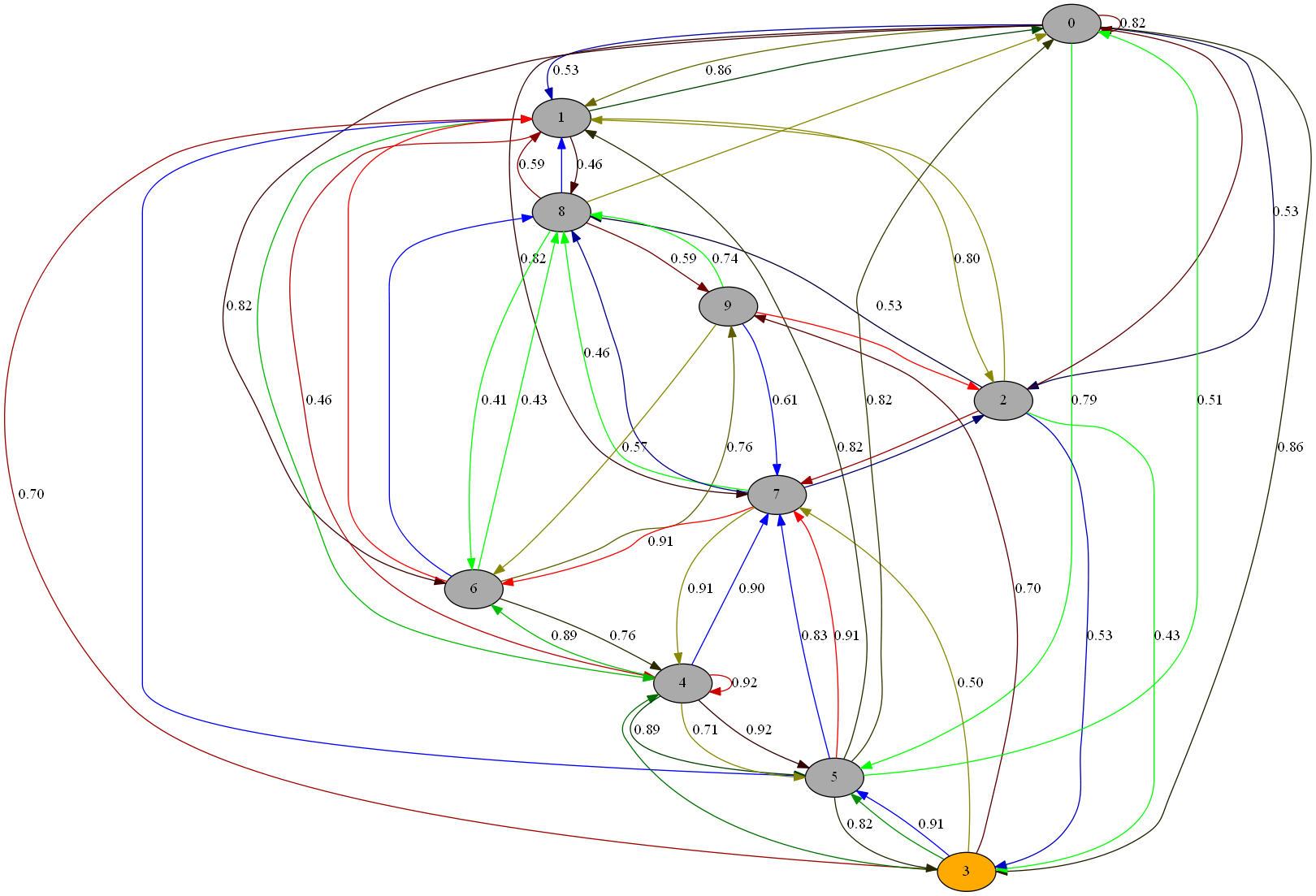}
	\end{center}
	\vspace{-5mm}
	\caption{A randomly-generated reward-rich MDP with $10$ states: One color per action, shaded according to the corresponding probability mass, labels indicate mean reward, and the current state is highlighted in orange.}\label{fig:randomrichMDP}
\end{figure}

The performance of \PSRL\ is good in the $6$-state RiverSwim environment, but degrades with a larger number of states when compared to \UCRLnew.
The performance of \PSRL\ in 2-room and 4-room MDPs are striking. These environments are goal-state (a.k.a.~goal-oriented) MDPs with very sparse rewards.
We conjecture  \PSRL\ favors such environment. In a Garnet MDP, we observe that \PSRL\ is not necessarily competitive.
 Figure~\ref{fig:PSRLrandomMDP}, left, shows the results in a $100$-state random MDP with relatively sparse rewards. \PSRL\ has a competitve initial phase, but is later outperformed by \UCRLnew, which suggests it is unable to find an optimal policy when $T$ is not very large. Figure~\ref{fig:PSRLrandomMDP}, right, shows the result of an experiment in a small $10$-state Garnet MDP,  but where most rewards are constrained to be far from $0$. Here, we observe that \PSRL\ achieves a very poor performance in such a case. 
 The MDP is depicted in Figure~\ref{fig:randomrichMDP} for completeness.

\end{document}

%% file: definitions.tex
\newcommand{\BEAS}{\begin{eqnarray*}}
\newcommand{\EEAS}{\end{eqnarray*}}
\newcommand{\BEA}{\begin{eqnarray}}
\newcommand{\EEA}{\end{eqnarray}}
\newcommand{\BEQ}{\begin{equation}}
\newcommand{\EEQ}{\end{equation}}
\newcommand{\BIT}{\begin{itemize}}
\newcommand{\EIT}{\end{itemize}}
\newcommand{\BNUM}{\begin{enumerate}}
\newcommand{\ENUM}{\end{enumerate}}
\newcommand{\beq}{\begin{equation}}
\newcommand{\eeq}{\end{equation}}
\newcommand{\beqa}{\begin{eqnarray}}
\newcommand{\eeqa}{\end{eqnarray}}
\newcommand{\beqan}{\begin{eqnarray*}}
\newcommand{\eeqan}{\end{eqnarray*}}
\newcommand{\bealn}{\begin{align*}}
\newcommand{\eealn}{\end{align*}}
\newcommand{\al}[1]{ \begin{align} #1  \end{align}}

\newcommand{\als}[1]{ \begin{align*} #1  \end{align*}}

\newcommand{\sk}{\nonumber\\}

\newcommand{\el}{\end{flushleft}}
\newcommand{\bl}{\begin{flushleft}}

\newcommand{\argmax}{\mathop{\mathrm{argmax}}}
\newcommand{\Argmax}{\mathop{\mathrm{Argmax}}}

\newcommand{\Argmin}{\mathop{\mathrm{Argmin}}}

\newcommand{\bA}{\mathbb{A}}

\newcommand{\bI}{\mathbb{I}}

\newcommand{\bS}{\mathbb{S}}

\newcommand{\cA}{\mathcal{A}}
\newcommand{\cB}{\mathcal{B}}
\newcommand{\cC}{\mathcal{C}}
\newcommand{\cD}{\mathcal{D}}

\newcommand{\cF}{\mathcal{F}}

\newcommand{\cK}{\mathcal{K}}

\newcommand{\cM}{\mathcal{M}}
\newcommand{\cN}{\mathcal{N}}
\newcommand{\cO}{\mathcal{O}}

\newcommand{\cS}{\mathcal{S}}

\newcommand{\cX}{\mathcal{X}}

\newcommand{\kR}{\mathfrak{R}}

\newcommand{\Real}{\mathbb{R}}
\newcommand{\Nat}{\mathbb{N}}

\newcommand{\indic}[1]{\mathbb{I}\{#1\}}

\renewcommand{\phi}{\varphi}
\renewcommand{\epsilon}{\varepsilon}

\def\NN{{\mathbb N}}
\def\EE{{\mathbb E}}
\def\PP{{\mathbb P}}
\def\RR{{\mathbb R}}

\def\Ocal{{\mathcal O}}

\newcommand{\Esp}{\mathbb{E}}
\newcommand{\Var}{\mathbb{V}}

\renewcommand{\Pr}{\mathbb{P}}

\newcommand{\supp}{\mathrm{supp}}

\newcommand{\kl}{\texttt{kl}}

%% file: RS_ICML2020.pdf_tex
\begingroup%
  \makeatletter%
  \providecommand\color[2][]{%
    \errmessage{(Inkscape) Color is used for the text in Inkscape, but the package 'color.sty' is not loaded}%
    \renewcommand\color[2][]{}%
  }%
  \providecommand\transparent[1]{%
    \errmessage{(Inkscape) Transparency is used (non-zero) for the text in Inkscape, but the package 'transparent.sty' is not loaded}%
    \renewcommand\transparent[1]{}%
  }%
  \providecommand\rotatebox[2]{#2}%
  \ifx\svgwidth\undefined%
    \setlength{\unitlength}{604.73983645bp}%
    \ifx\svgscale\undefined%
      \relax%
    \else%
      \setlength{\unitlength}{\unitlength * \real{\svgscale}}%
    \fi%
  \else%
    \setlength{\unitlength}{\svgwidth}%
  \fi%
  \global\let\svgwidth\undefined%
  \global\let\svgscale\undefined%
  \makeatother%
  \begin{picture}(1,0.23520102)%
    \put(0,0){\includegraphics[width=\unitlength,page=1]{RS_ICML2020.pdf}}%
    \put(0.93415466,0.12094756){\color[rgb]{0,0,0}\makebox(0,0)[lt]{\begin{minipage}{0.21134427\unitlength}\raggedright $s_L$\end{minipage}}}%
    \put(0.75026781,0.12037184){\color[rgb]{0,0,0}\makebox(0,0)[lt]{\begin{minipage}{0.33040373\unitlength}\raggedright $s_{L-1}$\end{minipage}}}%
    \put(0.93392041,0.23957471){\color[rgb]{0,0,0}\makebox(0,0)[lt]{\begin{minipage}{0.21134427\unitlength}\raggedright $0.6$\\ $(r=1)$\end{minipage}}}%
    \put(0.7594562,0.21904991){\color[rgb]{0,0,0}\makebox(0,0)[lt]{\begin{minipage}{0.21134427\unitlength}\raggedright $0.6$\end{minipage}}}%
    \put(0.84203212,0.15378257){\color[rgb]{0,0,0}\makebox(0,0)[lt]{\begin{minipage}{0.21134427\unitlength}\raggedright $0.35$\end{minipage}}}%
    \put(0.84160615,0.0520501){\color[rgb]{0,0,0}\makebox(0,0)[lt]{\begin{minipage}{0.21134427\unitlength}\raggedright $1$\end{minipage}}}%
    \put(0,0){\includegraphics[width=\unitlength,page=2]{RS_ICML2020.pdf}}%
    \put(0.67499125,0.15789925){\color[rgb]{0,0,0}\makebox(0,0)[lt]{\begin{minipage}{0.21134427\unitlength}\raggedright $0.35$\end{minipage}}}%
    \put(0.67494326,0.10360153){\color[rgb]{0,0,0}\makebox(0,0)[lt]{\begin{minipage}{0.21134427\unitlength}\raggedright $0.05$\end{minipage}}}%
    \put(0.67191952,0.05352102){\color[rgb]{0,0,0}\makebox(0,0)[lt]{\begin{minipage}{0.21134427\unitlength}\raggedright $1$\end{minipage}}}%
    \put(0,0){\includegraphics[width=\unitlength,page=3]{RS_ICML2020.pdf}}%
    \put(0.84198414,0.09787847){\color[rgb]{0,0,0}\makebox(0,0)[lt]{\begin{minipage}{0.21134427\unitlength}\raggedright $0.4$\end{minipage}}}%
    \put(0,0){\includegraphics[width=\unitlength,page=4]{RS_ICML2020.pdf}}%
    \put(0.02976956,0.12452384){\color[rgb]{0,0,0}\makebox(0,0)[lt]{\begin{minipage}{0.33040373\unitlength}\raggedright $s_1$\end{minipage}}}%
    \put(0.02308335,0.22320191){\color[rgb]{0,0,0}\makebox(0,0)[lt]{\begin{minipage}{0.21134427\unitlength}\raggedright $0.4$\end{minipage}}}%
    \put(0.10565928,0.15793457){\color[rgb]{0,0,0}\makebox(0,0)[lt]{\begin{minipage}{0.21134427\unitlength}\raggedright $0.6$\end{minipage}}}%
    \put(0.10561128,0.10099108){\color[rgb]{0,0,0}\makebox(0,0)[lt]{\begin{minipage}{0.21134427\unitlength}\raggedright $0.05$\end{minipage}}}%
    \put(0.10258754,0.0562021){\color[rgb]{0,0,0}\makebox(0,0)[lt]{\begin{minipage}{0.21134427\unitlength}\raggedright $1$\end{minipage}}}%
    \put(0,0){\includegraphics[width=\unitlength,page=5]{RS_ICML2020.pdf}}%
    \put(0.19371674,0.22480829){\color[rgb]{0,0,0}\makebox(0,0)[lt]{\begin{minipage}{0.21134427\unitlength}\raggedright $0.6$\end{minipage}}}%
    \put(-0.0012515,0.04109531){\color[rgb]{0,0,0}\makebox(0,0)[lt]{\begin{minipage}{0.21134427\unitlength}\raggedright $1$\\ $(r=0.05)$\end{minipage}}}%
    \put(0.19908608,0.12301893){\color[rgb]{0,0,0}\makebox(0,0)[lt]{\begin{minipage}{0.33040373\unitlength}\raggedright $s_2$\end{minipage}}}%
    \put(0,0){\includegraphics[width=\unitlength,page=6]{RS_ICML2020.pdf}}%
    \put(0.2793978,0.15899288){\color[rgb]{0,0,0}\makebox(0,0)[lt]{\begin{minipage}{0.21134427\unitlength}\raggedright $0.35$\end{minipage}}}%
    \put(0.27934981,0.10204939){\color[rgb]{0,0,0}\makebox(0,0)[lt]{\begin{minipage}{0.21134427\unitlength}\raggedright $0.05$\end{minipage}}}%
    \put(0.27897183,0.05726041){\color[rgb]{0,0,0}\makebox(0,0)[lt]{\begin{minipage}{0.21134427\unitlength}\raggedright $1$\end{minipage}}}%
    \put(0,0){\includegraphics[width=\unitlength,page=7]{RS_ICML2020.pdf}}%
    \put(0.37520148,0.12309989){\color[rgb]{0,0,0}\makebox(0,0)[lt]{\begin{minipage}{0.33040373\unitlength}\raggedright $s_3$\end{minipage}}}%
    \put(0.36851528,0.22177796){\color[rgb]{0,0,0}\makebox(0,0)[lt]{\begin{minipage}{0.21134427\unitlength}\raggedright $0.6$\end{minipage}}}%
    \put(0,0){\includegraphics[width=\unitlength,page=8]{RS_ICML2020.pdf}}%
    \put(0.45393872,0.15912398){\color[rgb]{0,0,0}\makebox(0,0)[lt]{\begin{minipage}{0.21134427\unitlength}\raggedright $0.35$\end{minipage}}}%
    \put(0.45389073,0.10218048){\color[rgb]{0,0,0}\makebox(0,0)[lt]{\begin{minipage}{0.21134427\unitlength}\raggedright $0.05$\end{minipage}}}%
    \put(0.45351275,0.05739151){\color[rgb]{0,0,0}\makebox(0,0)[lt]{\begin{minipage}{0.21134427\unitlength}\raggedright $1$\end{minipage}}}%
    \put(0,0){\includegraphics[width=\unitlength,page=9]{RS_ICML2020.pdf}}%
  \end{picture}%
\endgroup%